\documentclass[runningheads]{llncs}

\usepackage{eccv}

\usepackage{eccvabbrv}

\usepackage{graphicx}
\usepackage{booktabs}

\usepackage[accsupp]{axessibility}  

\usepackage[table]{xcolor}
\usepackage{graphicx} 
\usepackage{booktabs} 
\usepackage{array}    
\usepackage{tabularx} 
\usepackage{makecell}
\usepackage{tikz}
\usepackage{multirow}
\usepackage{pifont}

\input{preamble}

\usepackage{hyperref}

\usepackage{orcidlink}

\begin{document}

\title{Mismatch Quest: Visual and Textual Feedback for Image-Text Misalignment}

\titlerunning{Mismatch Quest}

\author{
Brian Gordon$^*$\inst{1,2}\orcidlink{0000-0002-3016-3690} \and
Yonatan Bitton$^*$\inst{2}\orcidlink{0000-0002-1185-6838} \and
Yonatan Shafir\inst{1,2}\orcidlink{0009-0007-5068-9124} \and
Roopal Garg\inst{2}\orcidlink{0009-0001-3343-9226} \and
Xi Chen\inst{2}\orcidlink{0000-0002-1581-4627} \and  
Dani Lischinski\inst{2,3}\orcidlink{0000-0002-6191-0361} \and 
Daniel Cohen-Or\inst{1,2}\orcidlink{0000-0001-6777-7445} \and 
Idan Szpektor\inst{2}\orcidlink{0009-0002-2746-4027}
}

\authorrunning{B.~Gordon et al.}

\institute{Tel Aviv University$^1$, 
Google Research$^2$ \\ The Hebrew University of Jerusalem$^3$}


\def\thefootnote{*}\footnotetext{Equal contribution}\def\thefootnote{\arabic{footnote}}

\maketitle

\vspace{-15pt}
\begin{abstract}
While existing image-text alignment models reach high quality binary assessments, they fall short of pinpointing the exact source of misalignment.
In this paper, we present a method to provide detailed textual and visual explanation of detected misalignments between text-image pairs. 
We leverage large language models and visual grounding models to automatically construct a training set that holds plausible misaligned captions for a given image and corresponding textual explanations and visual indicators. We also publish a new human curated test set comprising ground-truth textual and visual misalignment annotations. Empirical results show that fine-tuning vision language models on our training set enables them to articulate misalignments and visually indicate them within images, outperforming strong baselines both on the binary alignment classification and the explanation generation tasks.  Our code and human curated test set are available at: \url{https://github.com/MismatchQuest/MismatchQuest}. 
\end{abstract}
\vspace{-30pt}
\section{Introduction}
\label{sec:intro}

Recently, text/image generative models~\cite{dalle2,dalle_3, Parti, rombach2021highresolution, Imagen, ho2021classifierfree, Cho2020XLXMERT, PaLI} achieved remarkable capabilities. However, they still often generate outputs that are not semantically-aligned to the input, both for text-to-image (T2I) and image captioning~\cite{marcus2022preliminary, learning_compose}. They especially struggle with complex, nuanced, or out-of-distribution descriptions and fail to generate images which follow the prompt precisely~\cite{dalle2_see_double, attend_and_excite}.
As long as alignment quality is insufficient, adoption of Vision-Language Models (VLMs) may be limited.

\begin{figure}
  \centering
  \includegraphics[width=\columnwidth]{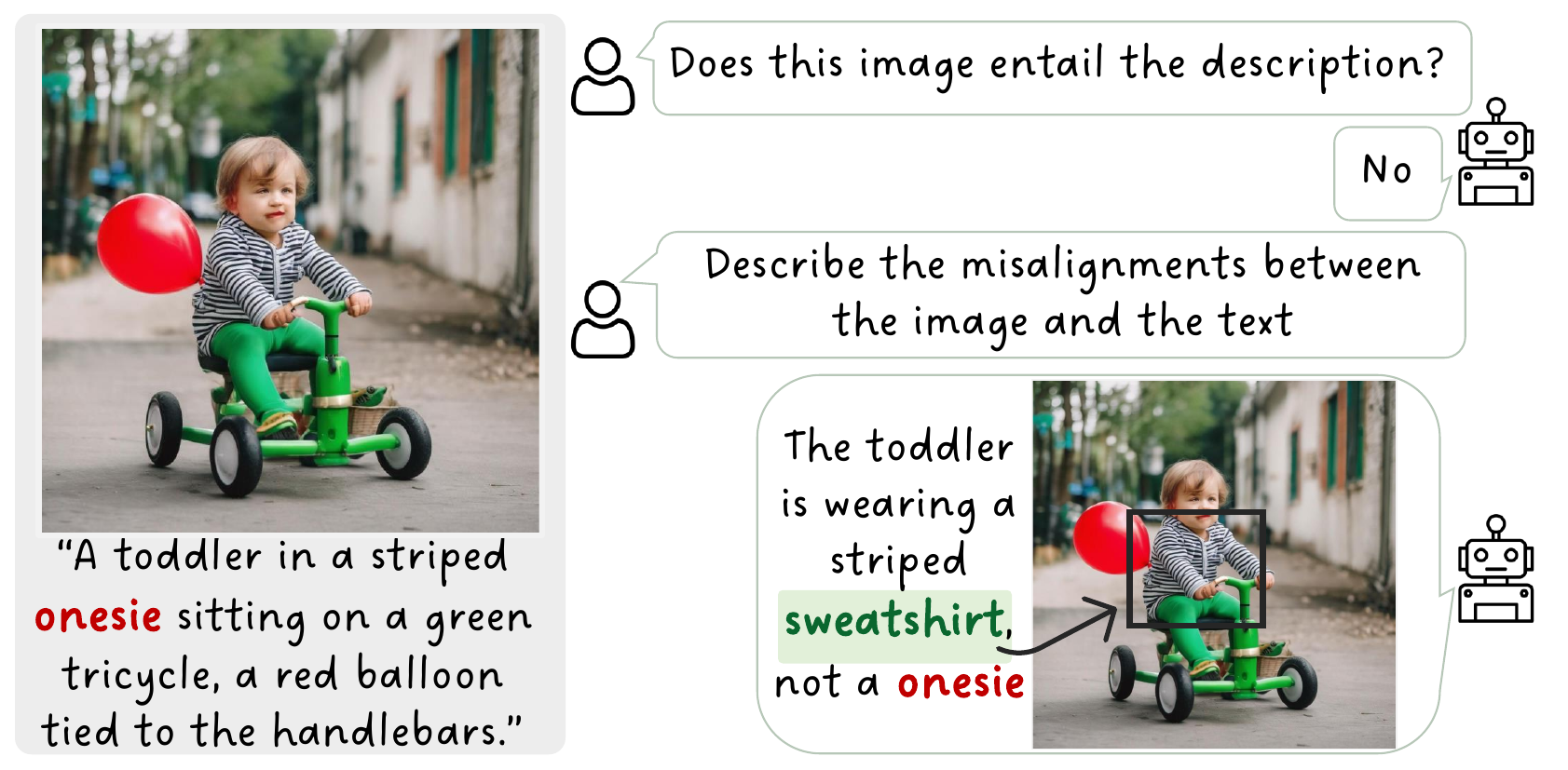}
\caption{Our alignment model steps: (1) the model predicts the alignment label between the input image/text pairs; (2) for misalignment labels, it then generates textual and visual feedback.}

  \label{fig:teaser}
\vspace{-17pt}
\end{figure}

\begin{figure*}
  \centering 
\includegraphics[width=\textwidth]{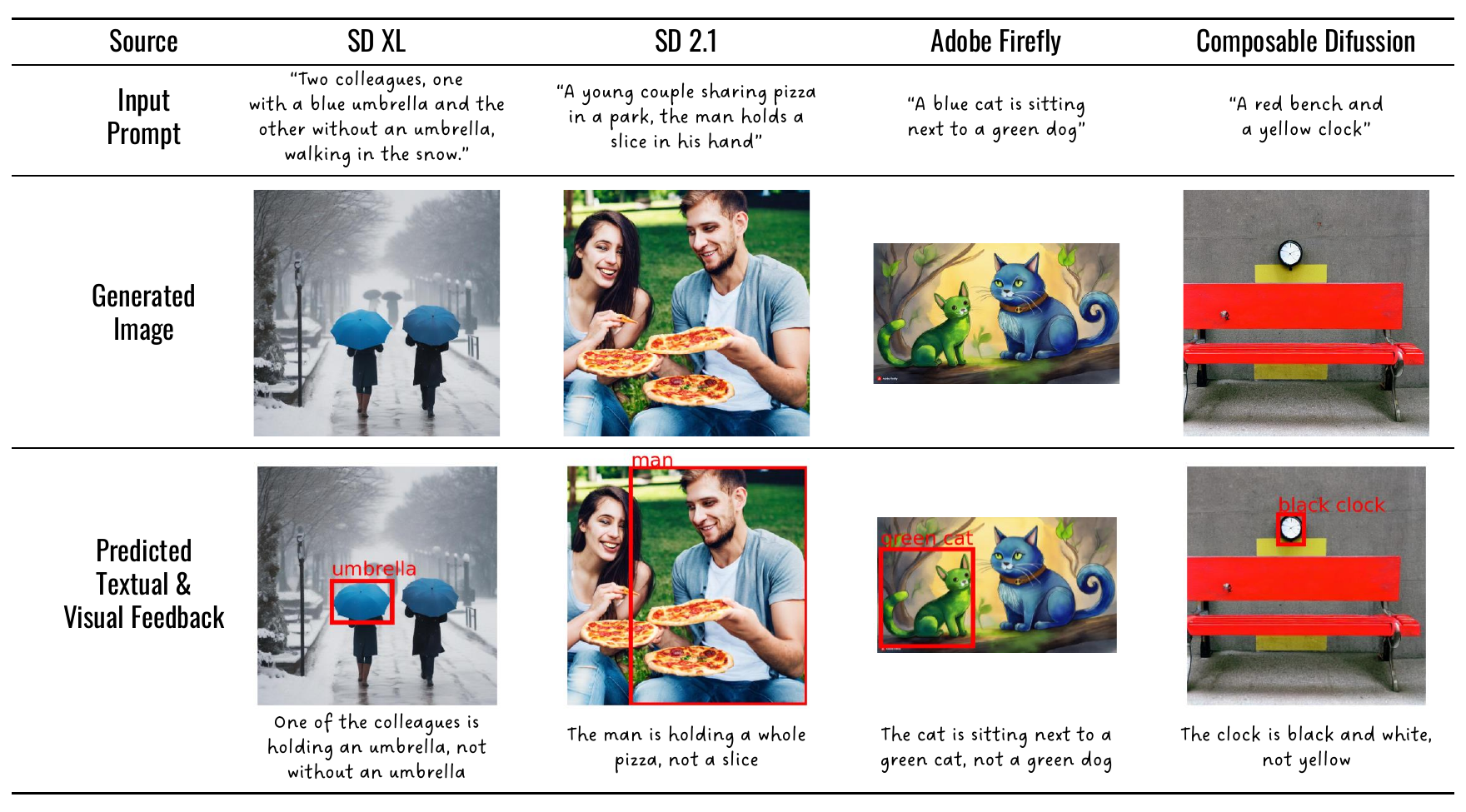}

\caption{Qualitative analysis of out-of-distribution results: Showcasing image-text pairs generated by Stable-Diffusion XL~\cite{podell2023sdxl_stable_difussion}, Stable-Diffusion 2.1~\cite{rombach2021highresolution}, Adobe Firefly~\cite{Adobe_Firefly} and Composable Diffusion~\cite{composable_diffusion} (credits to~\cite{attend_and_excite})  text-to-image models alongside the corresponding textual and visual feedback as predicted by the PaLI-X model finetuned on \trainset }
  \label{fig:OOD}
\vspace{-13pt}
\end{figure*}

To automatically gauge the alignment performance of VLMs, alignment evaluation models were proposed ~\cite{hessel2021clipscore, xu2023imagereward, WYSIWYR}. These models provide binary classification scores for text/image pairs. However, they do not offer insights regarding the misalignment: \emph{explanations} that could improve the understanding of VLM limitations and direct the training of better models.
To bridge this gap, we propose that alignment models should not only predict misalignments but also elucidate the specifics of text-image misalignments via both textual explanations and visual feedback using bounding boxes, as demonstrated in \Cref{fig:teaser} and \Cref{fig:OOD}.  We hypothesize that this novel form of feedback would deepen the understanding of misalignment causes within text/image pairs and facilitates the improvement of generative models.

\begin{figure*}
  \centering
  \includegraphics[width=\textwidth]{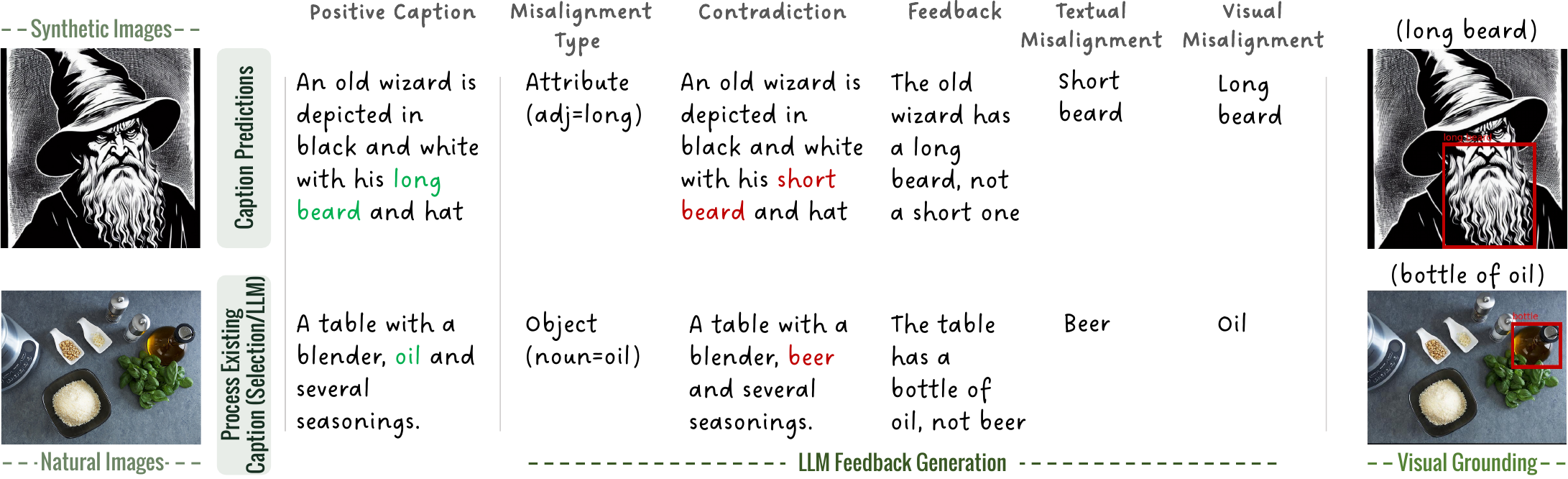}
\caption{The \method data generation method: Top image shows a synthetic image from PickaPic with a predicted caption; Bottom image is a natural image from COCO with its longest available caption. Both undergo LLM processing to generate contradictions, feedback, textual misalignment labels, and visual misalignment labels, followed by visual bounding box generation.}
  \label{fig:data_generation}
\vspace{-10pt}

\end{figure*}

To this end, we introduce \emph{\method}, a method that, for an aligned image/textual-caption pair, generates plausible contradicting captions on aspects such as entities, actions, attributes, and relationships, together with corresponding textual and visual (bounding-box) explanations of the misalignments (see \Cref{fig:data_generation}). This is done by employing the capabilities of large language models (LLMs) and visual grounding models. 
The outcome training set, denoted \textbf{T}extual and \textbf{V}isual (TV) \textbf{Feedback}, is a comprehensive compilation of 3 million instances, crafted to simulate a wide array of text-image scenarios from diverse databases including COCO~\cite{COCO}, Flickr30K~\cite{Plummer2015Flickr30kEC}, PickaPic~\cite{Kirstain2023PickaPicAO}, ImageReward~\cite{xu2023imagereward}, ADE20K~\cite{ade20k_zhou2017scene,ade20k_zhou2019semantic}, and OpenImages~\cite{openimages}.
We train an alignment evaluation model with this training set to both predict the alignment label and to generate feedback for misaligned image/text pairs.

To evaluate our alignment model, we construct and publish \emph{\testset}, a human-annotated test set. Human annotators provide textual explanations and approve visual bounding boxes to delineate misalignments, derived from a mixture of real and synthetic images and texts.
Our model outperforms other baselines across all metrics: including 10\% increase in alignment \emph{Accuracy}, 20\% increase in \emph{Entailment} w.r.t gold (human-annotated) textual feedback, and a 2-13\% increase in \emph{F1} for visual feedback. \emph{\testset} will be publicly available on our project page.
We complement our automated metrics with human ratings through an annotation study on Amazon Mechanical Turk~\cite{Amazon_Mechanical_Turk}, where our model outperforms the competing models by more than 100\% improvements on all metrics.
Our model also shows strong generalization capabilities with out-of-distribution images and prompts from various advanced T2I models such as Stable Diffusion (SD) v2.1~\cite{rombach2021highresolution}, SD XL~\cite{podell2023sdxl_stable_difussion}, Composable Diffusion~\cite{composable_diffusion}, and Adobe Firefly~\cite{Adobe_Firefly}. 
Finally, our ablation studies verify the advantage of our multitask training, a single model generating both misalignment labels and feedback for different prompts, compared to training individual models for each task, as well as the effectiveness of our training set filtering strategy. We aim to encourage future works based on the presented methodology for various cross-domain applications, such as enhancing text-to-image processes by providing a feedback signal. Furthermore, it can be utilized to identify and correct incorrect annotations in text-image pairs datasets and refine image captioning models by detecting erroneous captions.
In sum, our contributions are: (a) a feedback-centric data generation method (\method); (b) a comprehensive training set (\trainset); (c) a human-annotated evaluation set (\testset), which we also make publicly available; (d) trained models that surpass strong baselines. 

\section{Related Work}
\label{sec:related_work}

Our research intersects with developments in T2I generative models, vision-language models (VLMs), and approaches to T2I evaluation, emphasizing on automatic and explainable methods.

\textbf{Text-To-Image Generative Models.} T2I generation has evolved from Generative Adversarial Network (GAN) based models~\cite{gan_goodfellow, mansimov16_text2image, reed2016learning, pmlr-v48-reed16, Tao18attngan} to visual transformers and diffusion models, like DALL-E~\cite{dall_e,dalle2}, Parti~\cite{Parti}, Imagen~\cite{Imagen} and Stable Diffusion~\cite{rombach2021highresolution, podell2023sdxl_stable_difussion}.
While these models showcase improved capabilities in image generation from textual prompts, they still grapple with challenges in accurately reflecting intricate T2I correspondences~\cite{dalle2_see_double,Cho2022DallEval,petsiuk2022human}.

\textbf{Vision-Language Models.} LLMs like the GPT series ~\cite{gpt1,gpt2,gpt3,gpt4} have revolutionized various fields but primarily focus on text, limiting their efficacy in vision-language tasks. Recent advancements~\cite{recognition_to_cognition, PICa, yang2023mmreact, wu2023visual_chatgpt, instructblip, liu2023improvedllava, chen2023minigptv2, PaLI, chen2023pali3, Chen2023PaLIXOS, ye2023mplugowl, blip2, li2021grounded_glip} explore the synergy between visual components and LLMs to tackle tasks like image captioning and visual question answering (VQA), enhancing the understanding of visual content through textual descriptions.

\textbf{T2I Automatic Evaluation.} Traditional T2I evaluation methods utilize metrics like Fréchet Inception Distance (FID)~\cite{FID} and Inception Score~\cite{inception_score}. 
Alignment classification uses methods such as CLIP~\cite{clip}, CLIPScore~\cite{hessel2021clipscore}, and CLIP-R~\cite{CLIP_R}, or via image-captioning model comparison~\cite{papineni-etal-2002-bleu, anderson2016spice, CIDEr}. Methods such as~\cite{xu2023imagereward, Kirstain2023PickaPicAO} learn image quality reward models based on datasets with side-by-side human preferences and general ratings. In contrast, ~\cite{WYSIWYR} focuses on image-text alignment, producing alignment scores without detailed feedback on what is wrong with the generated image. Some studies~\cite{Cho2022DallEval, gokhale2022benchmarking_VISOR, hinz2019semantic} dissect alignment into components like object detection and color classification. Both datasets and automatic metrics lack detailed misalignment feedback, a gap that our work addresses.

\textbf{Image-Text Explainable Evaluation.} Recent studies, such as TIFA~\cite{hu2023tifa} and $VQ^2$~\cite{WYSIWYR}, offer an interpretable evaluation scheme by generating question-answer pairs from the text. These pairs are then analyzed using Visual Question Answering (VQA) on the image. DSG~\cite{JaeminCho2023_dsg} leverages this approach and creates a graph of questions, exploiting the dependencies between different questions and answers. These methods allow for detailed insights by contrasting expected text-based answers with image-derived responses, highlighting specific misalignments.

In a recent work, VPEval~\cite{Cho2023VPT2I_vpeval} generates a visual program using ChatGPT~\cite{openai-2022_chatgpt} and breaks down the evaluation process into a mixture of visual evaluation modules, which can be interpreted as an explanation. 

Our method aims for the direct generation of explanations for image/text discrepancies without the need for an interrogative question-answering pipeline or breaking the evaluation task into sub-tasks.

\section{Textual and Visual Feedback}
\label{sec:task}

\begin{table*}[!t]  
\caption{\trainset dataset examples including aligned and misaligned text-image pairs, and textual and visual misalignment feedback.}
\centering
\scriptsize 
\setlength{\tabcolsep}{3pt} 
\renewcommand{\arraystretch}{1} 
\renewcommand\tabularxcolumn[1]{m{#1}}

\begin{tabularx}{\textwidth}{@{} *{7}{>{\centering\arraybackslash}X} @{}} 
\toprule
\textbf{Source Dataset} & \textbf{PickaPic} & \textbf{ImageReward} & \textbf{COCO} & \textbf{Flickr30k} & \textbf{Open Images} & \textbf{ADE20K} \\
\midrule
\textbf{Images \& Texts} & Synthetic \& Synthetic & Synthetic \& Synthetic & Natural \& Natural & Natural \& Natural & Natural \& Synthetic & Natural \& Synthetic \\
\midrule
\textbf{\# Instances} & 1,982,362 & 56,392 & 418,653 & 37,327 & 577,717 & 19,825 \\
\midrule
\textbf{Image} & \includegraphics[width=\linewidth]{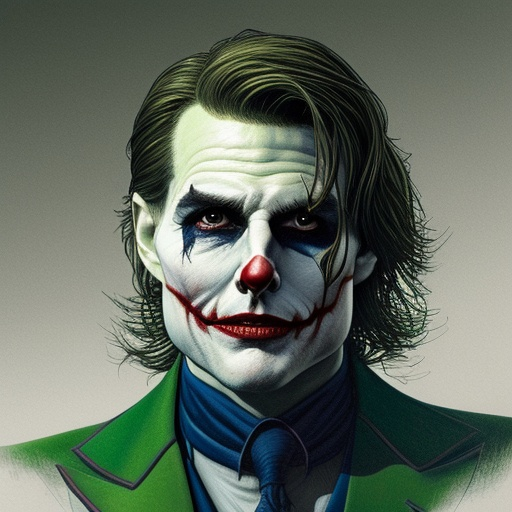} & \includegraphics[width=\linewidth]{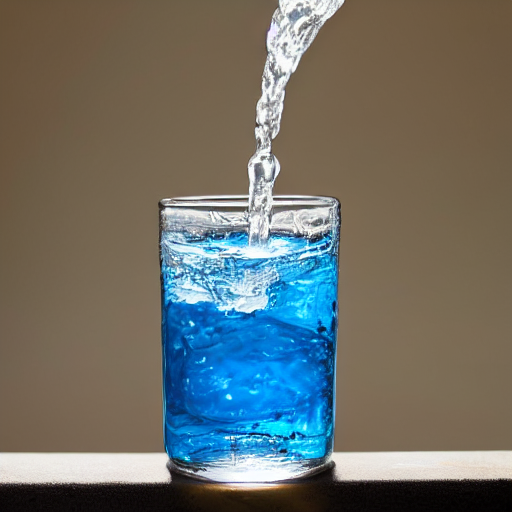} & \includegraphics[width=\linewidth]{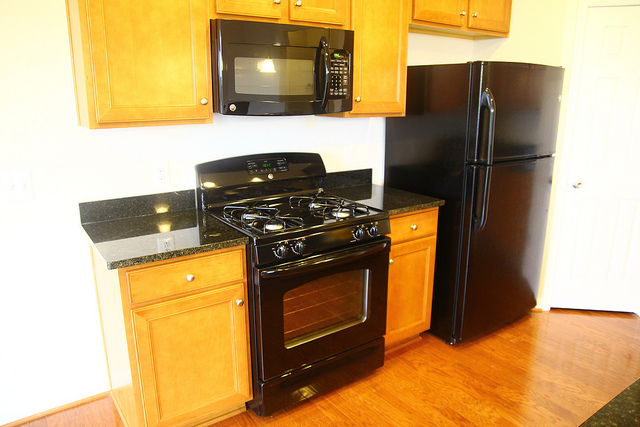} & \includegraphics[width=\linewidth]{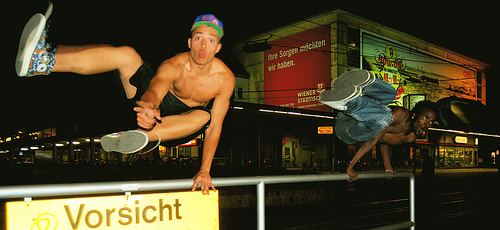} & \includegraphics[width=\linewidth]{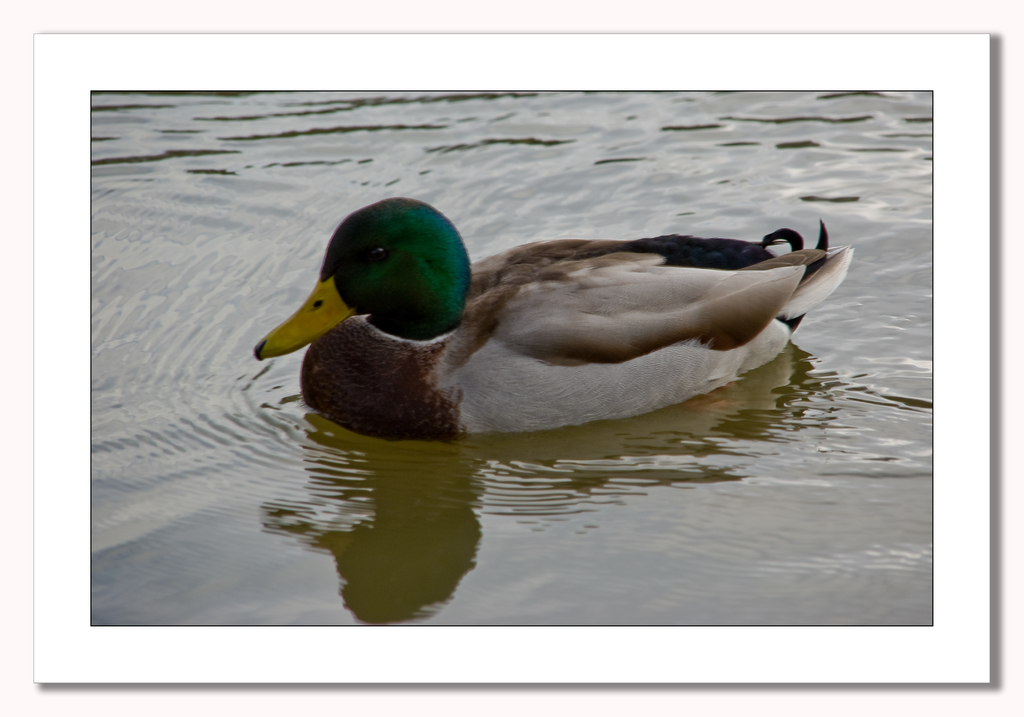} & \includegraphics[width=\linewidth]{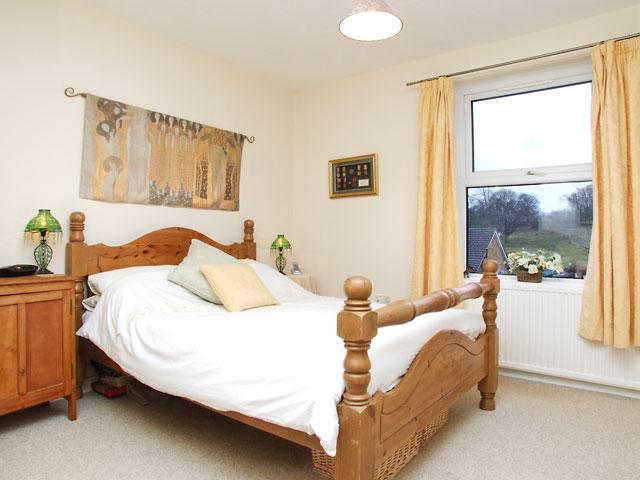} \\
\midrule
\textbf{Positive Caption} & A cartoon of a person dressed as a \textbf{joker} with a green coat and a blue tie against a gray background. & A close up of a glass of \textbf{blue} liquid on a table with a gray wall behind it . & A kitchen with cabinets, a stove, microwave and \textbf{refrigerator}. & Two men in Germany jumping \textbf{over} a rail at the same time without shirts. & A duck with a yellow beak is \textbf{swimming} in water. & A bed and a table with a lamp on it are in a room with a window and a view of \textbf{trees}. \\
\midrule
\textbf{Negative Caption} & A cartoon of a person dressed as a \textbf{clown} with a green coat and a blue tie against a gray background & A close up of a glass of \textbf{red} liquid on a table with a gray wall behind it. & A kitchen with cabinets, a stove, microwave and a \textbf{toaster}. & Two men in Germany jumping \textbf{under} a rail at the same time without shirts. & A duck with a yellow beak is \textbf{flying} in the air. & A bed and a table with a lamp on it are in a room with a window and a view of a \textbf{lake}. \\
\midrule
\textbf{Misalignment Type} & Object & Attribute & Object & Relation & Action & Object \\
\midrule
\textbf{Feedback} & The person is dressed as a joker, not a clown & The liquid is blue, not red & The kitchen is missing a toaster, but has a refrigerator. & The men are jumping over a rail, not under it & The duck is swimming, not flying & The room has a view of trees, not a lake \\
\midrule
\textbf{Misalignment in Text} & clown & red liquid & toaster & jumping under a rail & duck flying & a view of a lake \\
\midrule
\textbf{Visual Misalignment Detection} & [2, 3, 996, 995] joker & [380, 308, 944, 666] blue liquid & [193, 327, 347, 553] refrigerator & [277, 26, 664, 477] two men and [608, 3, 729, 998] a rail &[339, 245, 581, 834] duck swimming & [409, 727, 559, 930] trees \\
\bottomrule
\end{tabularx}

\label{table:train}
    
\end{table*}

Traditional image-text alignment evaluation models only provide alignment scores without detailed feedback. We propose to introduce a feedback mechanism, so that alignment models would not only score but also describe and visually annotate discrepancies between images and text.

In our multitask framework, as depicted in Figure \ref{fig:teaser}, a single model handles two main tasks. In the first task, \emph{Image-Text Entailment}~\cite{VNLI}, the model determines if an image corresponds to a given text description, outputs an alignment score to represent the likelihood of a ``yes'' answer\footnote{For direct comparison with other vision-language models, we present these outcomes as binary ``Yes/No'' responses instead of numerical scores.}.
The second task, \emph{Textual and Visual Feedback}, is performed when misalignments are detected in an input image-text pair. The model is expected to provide three outputs: (a) a textual summary of discrepancies between the pair; (b) identification of misaligned text segments; (c) image visual misalignments, marked by bounding boxes.

To equip a model with the tasks outlined above, we perform VLM fine-tuning. To this end, an extensive training set encompassing all necessary information is required. The primary challenge lies in creating a sufficiently large training set with suitable examples. The following section provides a detailed description of the methodology we employed for generating such a set.

\section{Training Dataset (TV Feedback) Generation} 
\label{sec:method}

\newcommand{\poscaption}{An old wizard is depicted in black and white with his long beard and hat.}
\newcommand{\negcaption}{An old wizard is depicted in black and white with his short beard and hat.}
\newcommand{\feedbacktext}{The difference between the captions is: the old wizard has a long beard, not a short one.}

To construct our training set, which is designed to detect and interpret misalignments in image-text pairs, we first collect aligned image-text pairs. Then, utilizing LLMs and visual grounding models, we generate negative examples with misalignments accompanied by textual and visual feedback (see examples in \Cref{table:train}). We next detail our approach, named \method.

\subsection{Collecting Positive Image-Text Pairs}
\label{subsec:positive_aggregation}

We compile a set of over a million positive image-text pairs, consisting of synthetic and natural images. Approximately 65\% of our examples consist of synthetic images, which were generated by a variety of T2I models from PickaPic~\cite{Kirstain2023PickaPicAO} and ImageReward~\cite{xu2023imagereward}. For these images, we employ the PaLI~\cite{PaLI} model to predict captions that are aligned with the image.

We also include natural images sourced from two well-established datasets, COCO~\cite{COCO} and Flickr30k~\cite{Plummer2015Flickr30kEC}. In these datasets, the images are already paired with human-annotated captions. When several captions are available per image, we select the longest to encourage textual richness. 

Finally, we take localized narratives~\cite{pont2020connecting}, captions offering a detailed point-of-view from the annotators) from ADE20k~\cite{ade20k_zhou2017scene,ade20k_zhou2019semantic} and OpenImages~\cite{openimages} and transform them into more conventional positive captions. To this end, we apply PaLM 2~\cite{anil2023palm} with a few-shot prompt (\ifAppendixEnable see \Cref{supmat_section:method_details}\else examples provided at the appendix\fi ) that rewrites the narratives into standardized captions.

\subsection{LLM Generation of Misaligned Image-Text Pairs and Feedback}
\label{subsec:congen_feedback}

For each positive example from \Cref{subsec:positive_aggregation} we derive negative examples that include misaligned captions and relevant feedback. This is a four step approach (\Cref{fig:data_generation}):

\textbf{1) Identify Misalignment Candidates.} 
For each aligned image/caption pair, we tag the caption for part of speech tags with spaCy~\cite{spacy}. We then define  four misalignment categories: object (noun), attribute (adjective), action (verb), and spatial relations. To ensure a balanced representation, we sample from these categories uniformly.

\textbf{2) Generate Misalignment and Textual Feedback.}
Per chosen misalignment candidate, we instruct PaLM 2~\cite{anil2023palm} API with few-shot prompts to automatically generate: (a) a contradiction caption that introduces the target misalignment; (b) a detailed explanation of the contradiction; (c) a misalignment cue that pinpoints the contradictory element in the caption; and (d) a label for the visual bounding box to be placed on the image. Our instructions and few-shot prompts are presented in \ifAppendixEnable \Cref{subsec:prompts} \else the appendix chapter\fi.

\textbf{3) Validate the Generation.}
Some LLM generations may be inaccurate. To increase the quality of the outputs, we filter out examples based on entailment validation as follows. Textual Entailment~\cite{dagan2010recognizing} models classify whether a \emph{hypothesis} text is entailed by a \emph{premise} text. We view this relationship as indicating the degree of semantic alignment. We use an entailment model by Honovich~\etal~\cite{q2} to assess the misalignment between our generated contradicting captions (hypothesis) and the original captions (premise), as well as the alignment between feedback (hypothesis) and caption (premise), as illustrated in \ifAppendixEnable \cref{sec:generation_validation} \else the appendix chapter\fi. Only valid contradictions and textual feedback, indicated by low and high entailment scores respectively, are retained.

\textbf{4) Annotate Visual Feedback.}
To create visual feedback for the target misalignment, we employ GroundingDINO~\cite{groundingdino}, which takes the textual label from PaLM 2's output and places a bounding box around the corresponding element in the image. To ensure consistent representation for different images, the bounding box coordinates are stored as a normalized range between 0 and 1000. 

To assess the quality of our \textbf{T}extual and \textbf{V}isual (TV)-\textbf{Feedback} training set, we sampled 300 generated items for manual inspection. The outcome of this rigorous human validation is a high confidence score of 91\%, which reflects the robustness of our automated generation process and the overall quality of the training dataset we have produced. 

\section{\testset Benchmark}
\label{sec:benchmark}

\begin{figure}
    \centering
    \includegraphics[width=0.7\columnwidth]{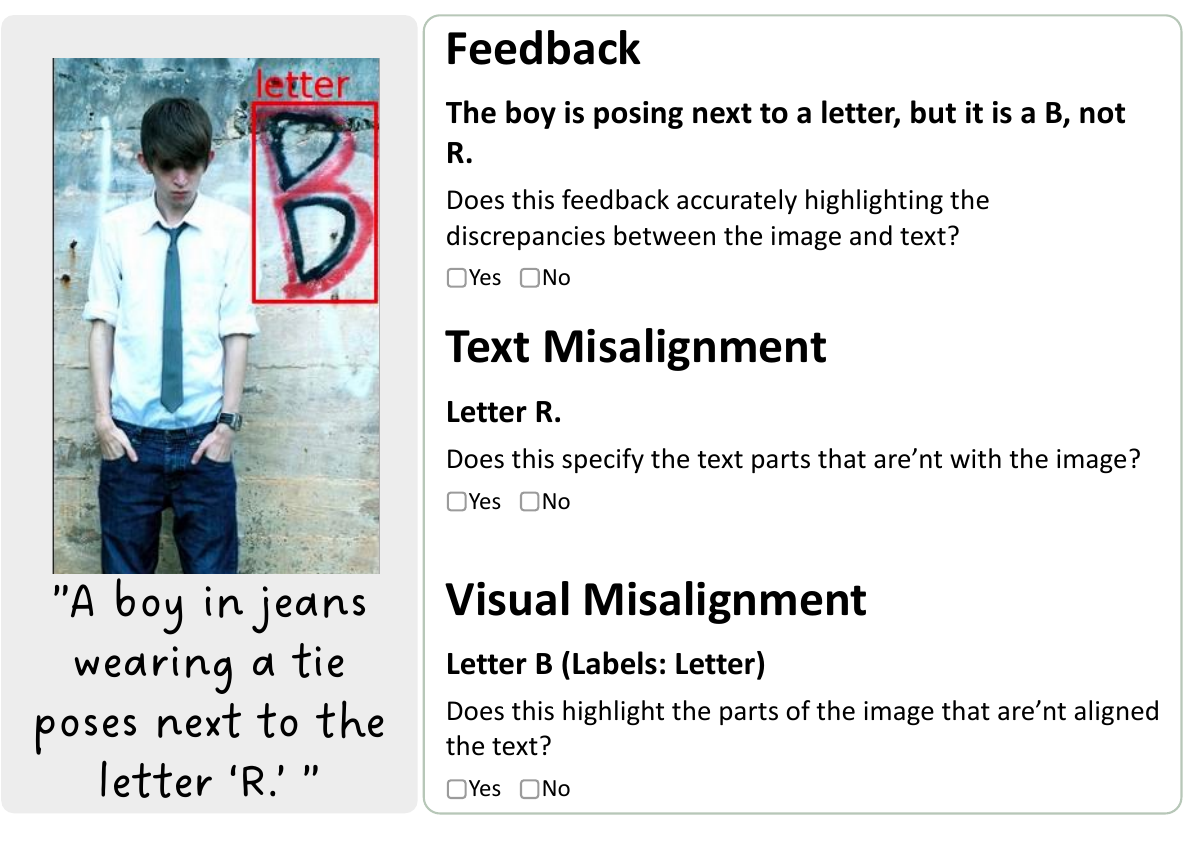}
    \caption{\testset annotation Amazon Mechanical Turk interface, questioning whether each part of the feedback, misalignment in text and misalignment in image are correct or not.}
    \label{fig:amt_example}
\vspace{-13pt}
\end{figure}

We present \testset, a comprehensive alignment benchmark. It features 2,008 human-annotated instances that highlight textual and visual feedback.

\subsection{Dataset Compilation}
\label{subsec:dataset_compilation}

The \testset Benchmark is based on the SeeTRUE dataset~\cite{WYSIWYR}, featuring aligned and misaligned image-text pairs. Each misaligned pair includes three human-generated descriptions detailing the misalignment. Similar to our method in Section \ref{sec:method}, we use PaLM 2 to generate a unified feedback statement at scale, covering both textual and visual misalignments. GroundingDINO then annotates these discrepancies on the images.

For verification, we conduct an annotation process on Amazon Mechanical Turk. Three annotators per instance, paid \$18 per hour, evaluated the accuracy of feedback and visual annotations (Figure \ref{fig:amt_example}). Only unanimously agreed instances, 66\% of the cases, were included in the final benchmark dataset.

\subsection{Evaluation Metrics}
\label{subsec:evaluation_metrics}
We compare alignment evaluation models on \testset using the following metrics:

\begin{itemize}
    \item \textbf{Image-Text Alignment:} 
    Binary Accuracy to gauge a model's ability to separate aligned and misaligned pairs.
  
    \item \textbf{Textual Feedback Quality:} 
    Using BART NLI \cite{lewis2019bart}\footnote{\url{huggingface.co/facebook/bart-large-mnli}}, we measure feedback quality by treating ground truth as the 'premise' and model predictions as the 'hypothesis', extracting an entailment score (0-1) as semantic alignment.    
    
    \item \textbf{Misalignment in Text:} This metric evaluates the model's ability to identify specific segments within the text that are not aligned with the corresponding image. Similar to the metric above, we use BART NLI to measure the entailment between the predicted text and the ground truth. The goal is to pinpoint the exact parts of the input text that are sources of misalignment.
    
    \item \textbf{Visual Misalignment Detection:} 
    We evaluate the model's bounding box generation using F1-Score@0.75 (indicating an IoU threshold of 0.75). This assessment combines precision and recall metrics to measure the accuracy of localization and object detection, ensuring a balance between avoiding missed objects (high precision) and minimizing false positives (high recall).
\end{itemize}

\begin{figure*}
\vspace{-13pt}
  \centering
    \includegraphics[width=\textwidth]{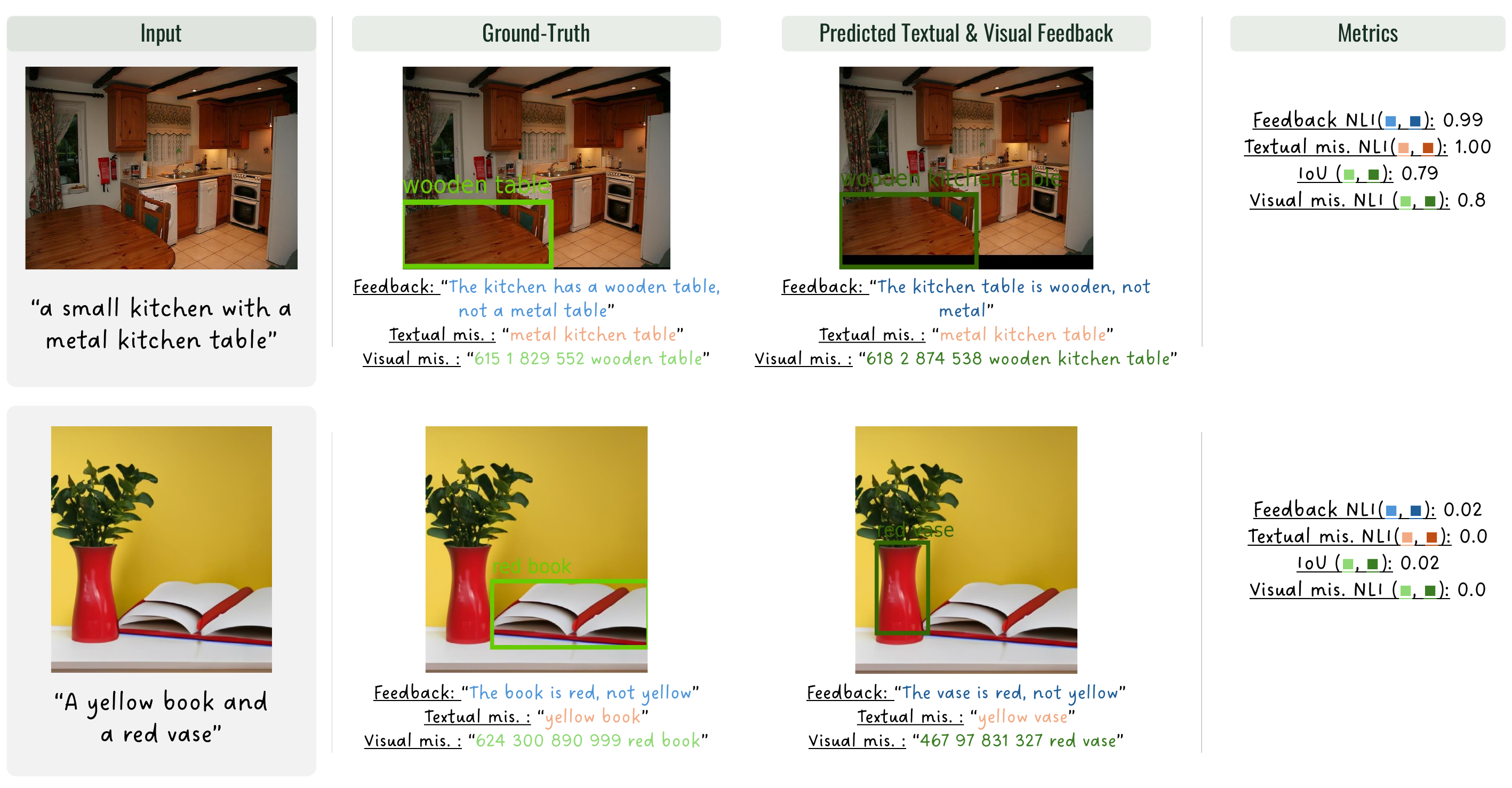}

\caption{Metric results on the \testset, showcasing calculations given the input, ground truth, and PaLI ft. model predictions, with NLI entailment scores calculated with BART NLI. The first row shows a high-scoring success example, while the second highlights a low-scoring failure with incorrect feedback and predictions.}

\label{fig:metrics_fig}
\vspace{-30pt}
\end{figure*}

We note that \emph{Image-Text Alignment} is applied to all 8,100 instances from the SeeTRUE dataset. The  other metrics are computed on \testset, containing only misaligned pairs. Examples showing our metric calculations can be seen at \Cref{fig:metrics_fig}.

\section{Experiments}
\label{sec:experiments}

This section describes our experiments, encompassing model selection, fine-tuning methods on \trainset, and thorough evaluation via the \testset benchmark. We also validate automated metric reliability through human annotation and assess model robustness with `out-of-distribution' examples from diverse sources.

\subsection{Models and Baselines}
\label{subsec:models}

Our experiments span multiple leading vision-language models, examined in both zero-shot and fine-tuned scenarios: MiniGPT-v2 (7B-ft)~\cite{chen2023minigpt}, LLaVa-1.5 (Vicuna-7b~\cite{liu2023improved}), InstructBLIP (FlanT5\textsubscript{XL})~\cite{dai2023instructblip}, mPLUG-Owl (LLaMa-7B-ft)~\cite{ye2023mplug}, PaLI Series~\cite{PaLI,chen2023pali3, Chen2023PaLIXOS}. Our methodology introduces a feedback task that, while new, aligns with the capabilities expected of leading VLMs, renowned for their instruction following capabilities. The task’s design mirrors scenarios these models encounter during training, ensuring they’re well-equipped to handle it.

For the zero-shot experiments, we queried the models with specific questions to assess their inherent capabilities:

\begin{enumerate}
    \item \textbf{Image-Text Entailment:} Assessing if an image semantically aligns with a given description (\textit{``Does this image entail \texttt{<text>}?''}).
       
    \item \textbf{Textual Misalignment Detection:} Identifying misaligned text elements (\textit{``Which part of \texttt{<text>} doesn't align with the image?''}).
        
    \item \textbf{Visual Misalignment Identification:}
    \textit{``What part of the following image is not aligned with the text: \texttt{<text>}?''} -- aimed at pinpointing visual discrepancies in the image relative to the text. 
\end{enumerate}

Our work uniquely offers an end-to-end assessment of both textual and visual misalignment. To evaluate baseline models for visual misalignment, we adopt a two-step approach. First, we ask for a textual misalignment description. Then, we employ the GroundingDINO grounding model to extract bounding-box information, since the baseline models do not output a bounding-box. {In addition, our fine-tuned model is capable of predicting the feedback along with both textual and visual misalignments in a single inference. To accurately assess our model's performance alongside the baselines, we report the visual misalignment performance using both the GroundingDINO output and our model's predictions

For the supervised experiments, we fine-tuned PaLI models with the visual question answering task using specific questions (additional fine-tuning details are in 
\ifAppendixEnable Appendix~\cref{sec:reproducibility} \else at the appendix \fi). The fine-tuning tasks encompass:

\begin{enumerate}
    \item \textbf{Image-Text Alignment:} Using the same query as in the zero-shot setup, \textit{``Does this image entail the description \texttt{<text>}?''}, we expected a binary `yes'/`no' response.

    \item \textbf{Textual and Visual Feedback:} We use a query for combined feedback: \textit{``Describe the misalignments between the image and the text: \texttt{<text>}''}. The expected response format is \textit{`<feedback> | <misalignment in text> | <misalignment in image (bounding-box)>'} , aiming to extract detailed feedback and specific misalignment indicators in a single model interaction.
\end{enumerate}

\begin{table*}
\caption{Comparative performance of image/text alignment models on the \testset Benchmark. ``ft.'' stands for fine-tuned on \trainset. Legend: $(*)$ marks the performance using PaLI bounding-box detector instead of GroundingDINO used for the baseline models.
}
    \centering
    \resizebox{0.95\textwidth}{!}{
    \begin{tabular}{lccccccc}
        \toprule
        & \multicolumn{2}{c}{\makecell{Feedback\\ NLI}} & \multicolumn{2}{c}{\makecell{Textual \\Misalignment NLI}} &  \multicolumn{2}{c}{\makecell{Visual Misalignment \\ F1-Score@0.75}} & \multicolumn{1}{c}{\makecell{Binary Class. \\ Acc.}} \\
        \cmidrule(lr){2-3} \cmidrule(lr){4-5} \cmidrule(lr){6-7} \cmidrule(lr){8-8}
        Model / Split & Test & Val & Test & Val & Test & Val & Test\\
        \midrule
        PaLI-3~\cite{chen2023pali3}
            & 0.18 & 0.22 & 0.23 & 0.46 & 0.47/0.47* &  0.35/0.48* & 0.51 \\
        InstructBLIP (FlanT5\textsubscript{XL}) \cite{instructblip}
            & 0.41 & 0.39 & 0.56 & 0.50 & 0.48 & 0.39 & 0.74 \\
        mPLUG-Owl (LLaMa-7B-ft) \cite{ye2023mplugowl}
            & 0.63 & 0.58 & 0.30 & 0.35 & 0.43 & 0.48 & 0.50 \\
        MiniGPT-v2 (7B-ft) \cite{chen2023minigptv2}
            & 0.46 & 0.37 & 0.56 & 0.58 & 0.44 & 0.43 & 0.68 \\
        LLaVa-1.5 (Vicuna-7b) \cite{liu2023improvedllava}
            & 0.57 & 0.48 & 0.17 & 0.21 & 0.43 & 0.48 & 0.72 \\
        \midrule
        PaLI-3 ft. Multitask~\cite{chen2023pali3}
            & 0.72 & \textbf{0.88} & 0.76 & \textbf{0.92} & 0.61/0.49* & 0.83/0.57* & 0.75 \\
        PaLI ft. Multitask~\cite{PaLI}
            & \textbf{0.75} & 0.87 & \textbf{0.78} & 0.92 & \textbf{0.65}/0.35* & \textbf{0.84}/0.39* & 0.77 \\
        PaLI-X ft. Multitask~\cite{Chen2023PaLIXOS}
            & 0.74 & 0.87 & 0.76 & 0.90 &  0.61/0.49* & 0.84/0.55* & \textbf{0.79} \\
        \bottomrule
    \end{tabular}}

    \label{tab:main_results}
\end{table*}

\begin{figure*}
\vspace{-8pt}
  \centering
    \includegraphics[width=\textwidth]{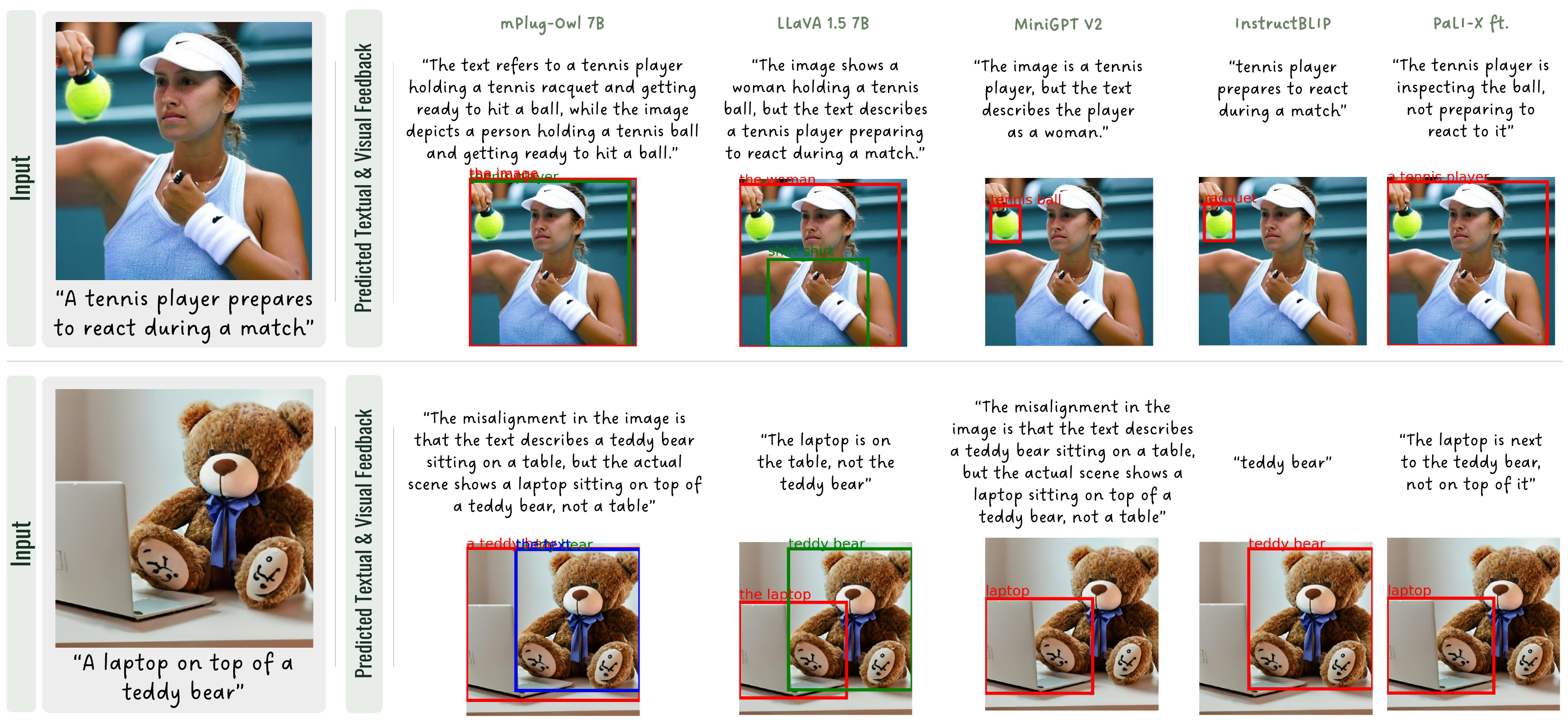}

\caption{Qualitative comparison of model outputs on two examples from \testset. The PaLI-X model, fine-tuned on \trainset, effectively identifies a distinct misalignment related to the tennis player's action and the relative position between the teddy bear and laptop, demonstrating its refined feedback ability.}
\label{fig:qualitative_test_results}
\vspace{-5pt}

\end{figure*}

\subsection{Main Results}
\Cref{tab:main_results} presents our main results on the \testset benchmark, and \Cref{fig:qualitative_test_results} provides qualitative examples. \emph{Val} results refer to ``in-distribution’’ auto-generated data, while \emph{Test} results  refer to ``out-of-distribution’’ human created examples.


Overall, the PaLI models fine tuned on \trainset outperform the baselines on all metrics. For example, Non-PaLI models achieved Feedback NLI scores from 0.406 to 0.627, while PaLI models reached 0.718 to 0.749. 
The largest, PaLI-X~\cite{Chen2023PaLIXOS} model achieved the highest performance on the binary alignment classification task. Surprisingly, it underperformed the smaller PaLI models on most feedback generation tasks. Specifically, the smaller but most recent PaLI-3 model, is best performing on the in-distribution testset, but less so on the out-of-distribution examples.
The PaLI models gap over the baselines is very large on the textual feedback tasks, but less so on the bounding box task. In future work, we plan to improve the multitasking efficiency of the fine-tuned models. \Cref{fig:metrics_fig} shows metrics results calculated on \testset examples to give a more clear overview. More details about  our metrics and evaluation process are available at \ifAppendixEnable \Cref{supmat_section:evaluation_metrics}\else the appendix chapter\fi.


\subsection{Human Ratings and Auto-Metrics Correlations}

For unbiased model evaluations and automatic metric validation, we conducted an Amazon Mechanical Turk study involving 1,500 instances. These instances included 250 samples from each of the six models used in our experiments. Annotators were assigned to evaluate the accuracy of these models in identifying and describing image-text misalignments, with each of the 1,500 instances being rated by three human raters.


\ifAppendixEnable\Cref{tab:results_human_rating} displays \else At the appendix chapter we present \fi results, highlighting PaLI-X with top scores in feedback accuracy (75.7\%), textual misalignment (80.1\%), and visual misalignment detection (63.5\%), showcasing superior alignment with human judgments. \ifAppendixEnable In \Cref{supmat_section:evaluation_metrics}, we present the annotators’ agreement chart for each model's predictions. \else We present the annotators’ agreement chart for each model's predictions as well. \fi

We evaluated our auto-evaluation metrics against 1,750 human ratings. Textual metrics included BART NLI~\cite{lewis2019bart}, BLEU-4~\cite{papineni2002bleu}, ROUGE-L~\cite{lin2004rouge}, METEOR~\cite{banerjee2005meteor}, CIDEr~\cite{vedantam2015cider}, BERTScore~\cite{zhang2019bertscore}, and TRUE NLI~\cite{honovich-etal-2022-true-evaluating}. Visual metrics comprised AP, IoU, Precision, Recall, and F1-Score at 0.75 threshold. \Cref{fig:correlations} shows the correlations, identifying BART NLI and F1-Score@0.75 as the most correlated textual and visual metrics, respectively. This analysis confirms the relevance and reliability of our automatic evaluation measures.

\begin{table}[ht]
\small
\caption{Human annotation results comparing model performances in feedback accuracy and misalignment identification. The values represent the mean percentage of ``yes'' responses from annotators. T. Misalignment stands for textual misalignments, and V. Misalignment for visual misalignments.}
\centering
{
\begin{tabular}{@{}lccc@{}}
\toprule
\textbf{Model}   &  \textbf{Feedback} & \textbf{T. Misalignment} & \textbf{V. Misalignment} \\ \midrule
PaLI-X ft. & \textbf{75.7} & \textbf{80.1} & \textbf{63.5}                        \\
PaLI-3 ft. & 68.1 & 72.4 & 61.6\\
LLaVA 1.5 7B & 29.9 & 5.1 & 16.2                        \\
mPlug-Owl 7B & 14.22 & 5.5 & 5.9\\
MiniGPT V2 & 11.6 & 39.1 & 21.7 \\
InstructBLIP & 1.3 & 32.6 & 29.9 \\ 
\bottomrule
\end{tabular}}

\label{tab:results_human_rating}
\vspace{-30pt}
\end{table}

\subsection{Out-of-distribution Generalization}
\label{sec:ood}
We evaluate our model's generalization capabilities on 100 `in-the-wild' Text-to-Image (T2I) generations from academic papers~\cite{attend_and_excite, rassin2023linguistic, saharia2022photorealistic} and Reddit, created using models like Adobe Firefly~\cite{Adobe_Firefly}, Composable Diffusion~\cite{composable_diffusion}, and Stable Diffusion versions 1.0 and 2.1. \Cref{fig:OOD} shows a selection of these results, with more available at \ifAppendixEnable \Cref{supmat_section:additional_results} \else the appendix chapter\fi .

We employed the fine-tuned PaLI-X model on \trainset to predict textual and visual feedback, and these results were rated by three human annotators following our benchmark protocol (Section \ref{sec:benchmark} and Figure \ref{fig:amt_example}).

Results indicated a feedback accuracy of 71\%, textual misalignment detection accuracy of 80\%, and visual misalignment accuracy of 60\%, showcasing the model's broad generalization to various out-of-distribution prompts and models. These findings also highlighted areas for potential model enhancement.

\begin{figure*}
    \centering

    \begin{subfigure}[b]{0.32\textwidth}
        \includegraphics[width=\textwidth]{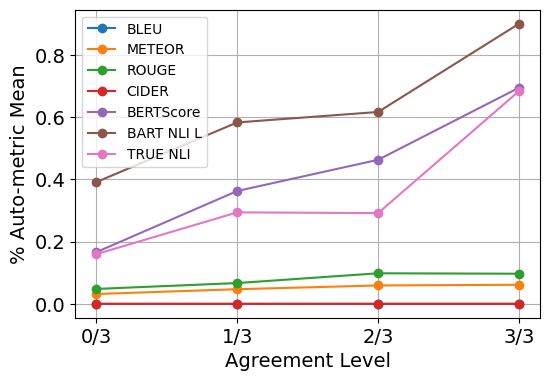}
        \caption{Feedback agreement}
        \label{fig:subfig1}
    \end{subfigure}
    \hfill
    \begin{subfigure}[b]{0.32\textwidth}
        \includegraphics[width=\textwidth]{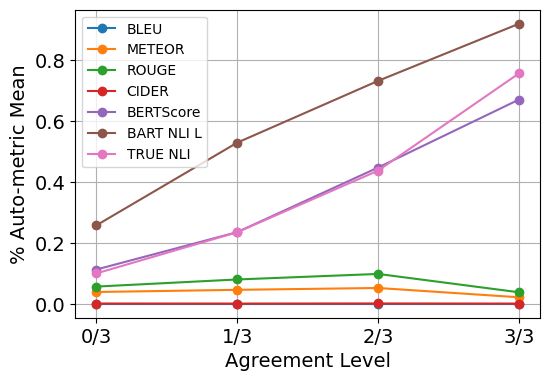}
        \caption{Textual agreement}
        \label{fig:subfig2}
    \end{subfigure}
    \hfill
    \begin{subfigure}[b]{0.32\textwidth}
        \includegraphics[width=\textwidth]{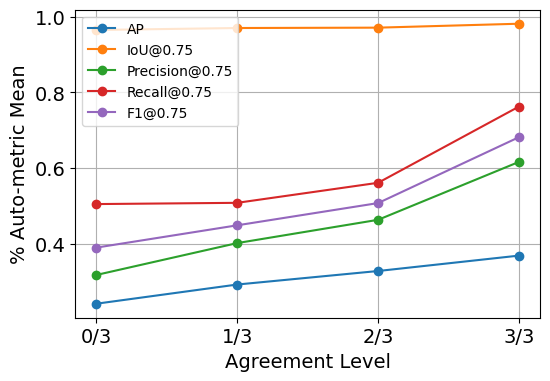}
        \caption{Visual agreement}
        \label{fig:subfig3}
    \end{subfigure}
    \caption{Correlation analysis between human ratings and automated metrics for feedback evaluation. Subfigures (a) and (b) explore textual feedback correlations with metrics like BART NLI and BERTScore, while subfigure (c) illustrates visual feedback correlations with metrics like IoU and F1-Score. The X-axis denotes annotator agreement; the Y-axis shows mean metric scores, identifying most correlated metrics.}

    \label{fig:correlations}
\vspace{-20pt}
\end{figure*}

\section{Analysis and Limitations}
\label{sec:analysis}

In this section, we analyze methodological ablation studies and discuss the limitations along with future directions for enhancing our model.

\subsection{Methodological Ablations Studies}
We conduct an ablation study to evaluate our methodologies. Our multi-task training approach achieves superior performance, with 75\% entailment accuracy and a 0.72 BART-NLI~\cite{lewis2019bart} score in feedback, highlighting its efficiency. Fine-tuning on our filtered dataset (77\% of total data) improves feedback and entailment tasks but degrades others, underscoring the positive impact of NLI model-based filtering. In a 2-step experiment simulating baselines, using GroundingDino~\cite{groundingdino} for grounding with predicted visual misalignment text labels improves bounding-box precision by 0.11 in F1-Score, showcasing its efficacy over our model.

\begin{table}[ht]
\vspace{-10pt}
\caption{Comparing PaLI-3~\cite{chen2023pali3} models: baseline, fine-tuned (entailment, feedback), multitask (unfiltered data, entailment+feedback). $+GD$ denotes two-step visual misalignment (b-box) prediction via GroundingDino. The study underscores the benefits of multitask training and the effectiveness of dataset filtering in enhancing performance.}
    \centering
    {
    \begin{tabular}{lccccc}
        \toprule
        Model & \makecell{Feedback \\ NLI} & \makecell{Textual Mis. \\  NLI} &  \makecell{Visual Mis. \\ F1@0.75} & \makecell{Binary \\ Acc.} \\
        \midrule
        Baseline
            & 0.18 & 0.23 & 0.47 & 0.51  \\
        Entailment
            & - & - & - & 0.74  \\
        Feedback
            & 0.70 & 0.76 & 0.50 & - &   \\
        Feedback+GD
            & 0.72 & 0.77 & \textbf{0.61} & - &   \\
        Multitask (Unf.)
            & 0.69 & \textbf{0.80} & 0.51 & 0.74   \\
        Multitask
            & \textbf{0.72} & 0.77 & 0.49 & \textbf{0.75}   \\
        \bottomrule
    \end{tabular}
    }

    \label{tab:ablation_results}
\vspace{-25pt}

\end{table}


\subsection{Limitations and Future Work}
In our evaluation across various datasets, our model showed proficiency but also revealed key improvement areas:

\begin{itemize}
    \item \textbf{No Visual Feedback:} In cases where no visual feedback is expected (\Cref{fig:limitations}a), our model incorrectly predicts it. To address this, we plan to enrich \trainset with scenarios like ``an image of a horse'' becoming ``an image of a horse \emph{and a dog},'' with feedback like ``there is only a horse, not a dog and a horse,'' and without generating a bounding box.
    \item \textbf{Multiple Misalignments:} Instances requiring identification of multiple misalignments (\Cref{fig:limitations}b). Our model often detects only one issue where several exist. We will enhance \trainset with cases like transforming ``a \emph{white dog} and a \emph{black cat}'' into ``a \emph{white cat} and a \emph{black dog},'' with feedback addressing both color and species misalignments and bounding boxes highlighting each.
    In \ifAppendixEnable \Cref{supmat_section:additional_results} \else the appendix chapter \fi , we show how a finetuned model detects multiple misalignments sequentially using a MagicBrush~\cite{Zhang2023MagicBrush} dataset example. The model identifies one misalignment at a time, and feedback signals guide an instruction editing model to iteratively correct them.

    \item \textbf{Loose Bounding Boxes:} As observed in \cref{fig:OOD} for the SD2.1 example, our model occasionally generates loose bounding boxes. For instance, rather than confining the b-box to the pizza, it may encompass the entire person.
\end{itemize}

These enhancements to the \trainset are aimed to improve the model's ability to address various misalignment types, making it more effective and applicable in real-world situations.
\begin{figure}
    \centering
    \includegraphics[width=0.9\columnwidth]{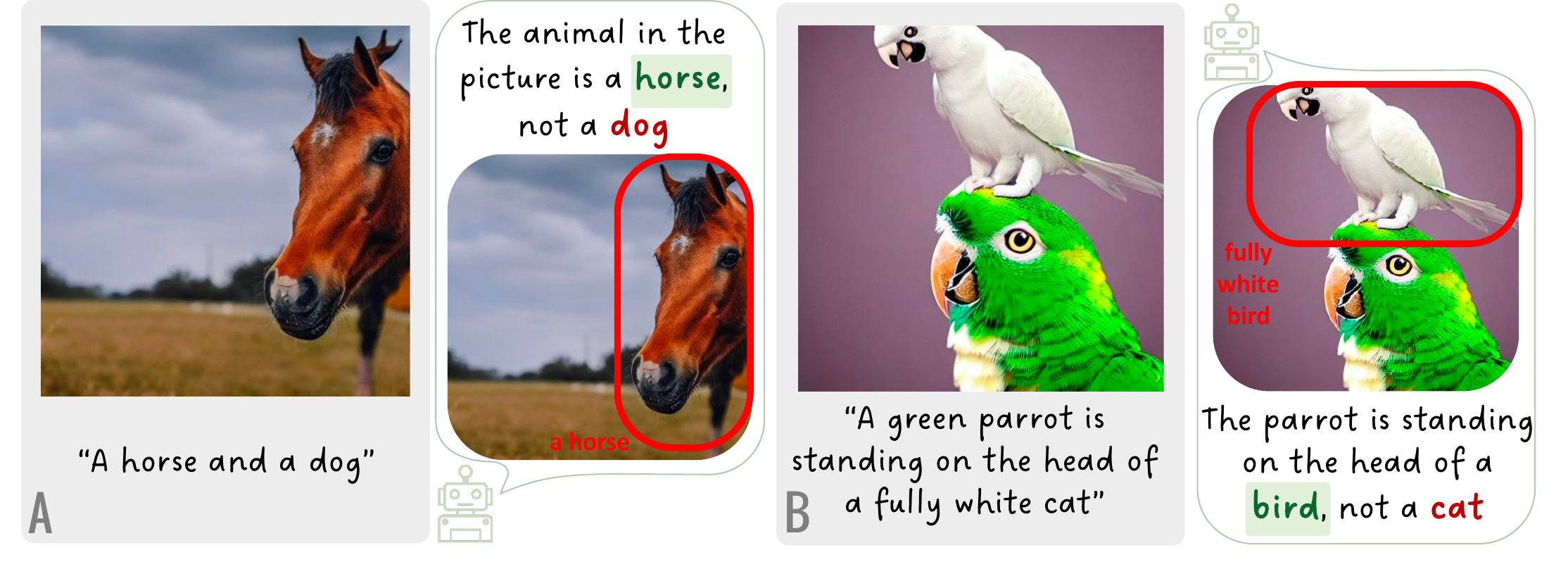}
\caption{Model limitations: (a) Misalignment due to a \emph{missing object}, where the model incorrectly adds a bounding box over a horse; (b) Multiple misalignments, with the model only identifying one - the top parrot should be green and the bottom a white cat. The model requires multiple iterations for full correction.}

        \label{fig:limitations}
\vspace{-30pt}
\end{figure}

\section{Conclusion}
\label{sec:conclusion}
Our research develops an end-to-end strategy providing visual and textual feedback for text-to-image models, targeting and clarifying alignment issues for refinement. We introduced \trainset, a specialized dataset for fine-tuning feedback in these models, leading to several robust developments. This dataset and methodology demonstrate broad potential, notably in enhancing text-to-image generation, improving dataset annotation accuracy, and refining image captioning through detailed feedback. Our comprehensive testing on \testset and various scenarios validates our approach's effectiveness. While primarily aimed at text-to-image feedback enhancement, we anticipate our work will significantly improve generative model accuracy across different domains.



%
%
\bibliographystyle{splncs04}
\bibliography{main}

\ifAppendixEnable 
    \appendix
    \clearpage
\setcounter{page}{1}
\begin{center}
 \textbf{\large Mismatch Quest: Visual and Textual Feedback for Image-Text Misalignment}\\
 \vspace{0.5em}{\large Supplementary Material}
\end{center}


\section{Additional Results}
\label{supmat_section:additional_results}
In \Cref{fig:supmat_competitors}, we present additional examples comparing our model-predicted Textual-Visual feedback against baseline competitors. We would like to emphasize that our PaLI fine-tuned models produce the full feedback end-to-end, including the bounding box. In contrast, for the other models, we are required to use the GroundingDino~\cite{groundingdino} model to complete a full feedback for them, as they don't have the ability to output bounding boxes.

Further qualitative results on our \testset are available at \Cref{fig:supmat_seetrue_qualitative}.

\Cref{fig:supmat_ood} presents another set of examples showcasing our model's performance in an `in-the-wild' setting using images generated by state-of-the-art text-to-image models: Stable-Diffusion XL~\cite{podell2023sdxl_stable_difussion}, StableDiffusion 2.1~\cite{rombach2021highresolution}, and Adobe Firefly~\cite{Adobe_Firefly}.

As illustrated in \cref{fig:serial_editing}, we show how a fine-tuned model can sequentially detect multiple misalignments using an example from the MagicBrush dataset. While the model identifies misalignments one at a time, the feedback signals direct an instruction editing model to iteratively correct these errors. This method can be used as a baseline for future research addressing multiple misalignments, particularly in the context of long image descriptions.

\begin{figure*}
  \centering
  \includegraphics[width=\textwidth]{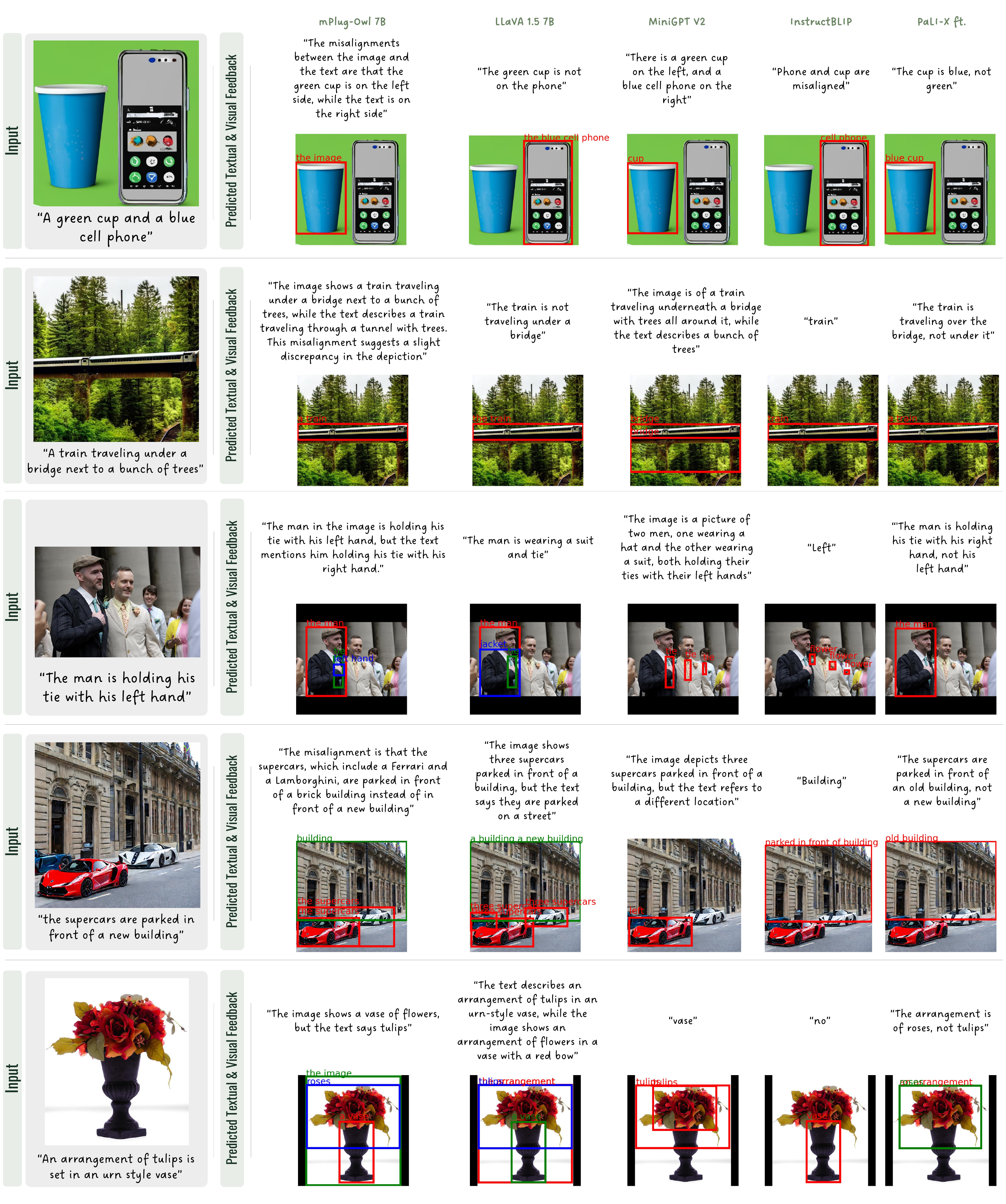}
\caption{Additional qualitative comparison examples from \testset. The PaLI-X  model, fine-tuned on \trainset, produces the most aligned and concise feedback compared to the rest of the models.}
  \label{fig:supmat_competitors}
\end{figure*}

\begin{figure*}
  \centering
  \includegraphics[width=\textwidth]{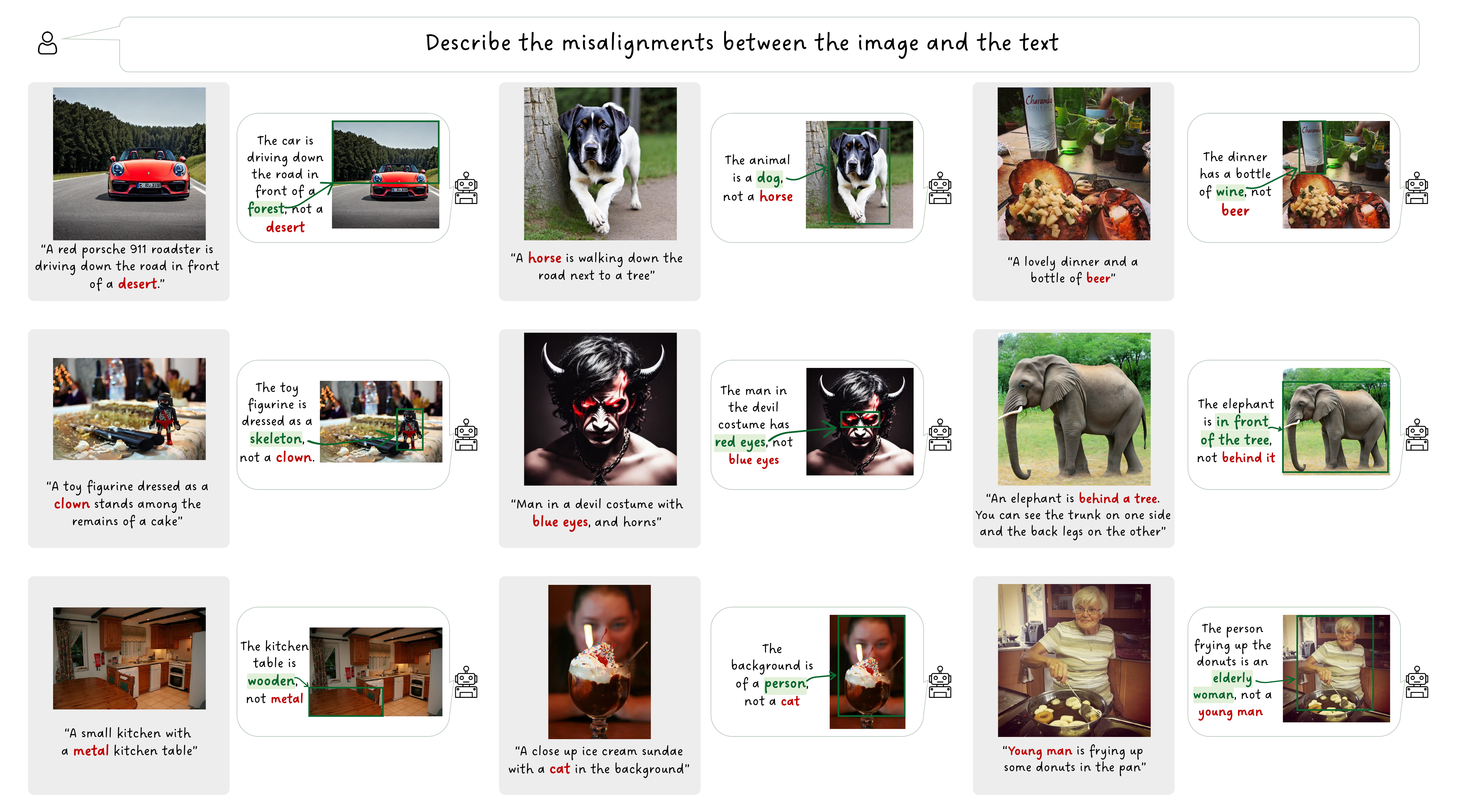}
\caption{Additional qualitative results on \testset. Please note the precision of our produced Textual-Visual feedback, which includes: a concise explanation of the misalignment, a misalignment cue that pinpoints the contradictory source in the caption, and a labeled bounding box.}
  \label{fig:supmat_seetrue_qualitative}
\end{figure*}

\begin{figure*}
  \centering
  \includegraphics[width=\textwidth]{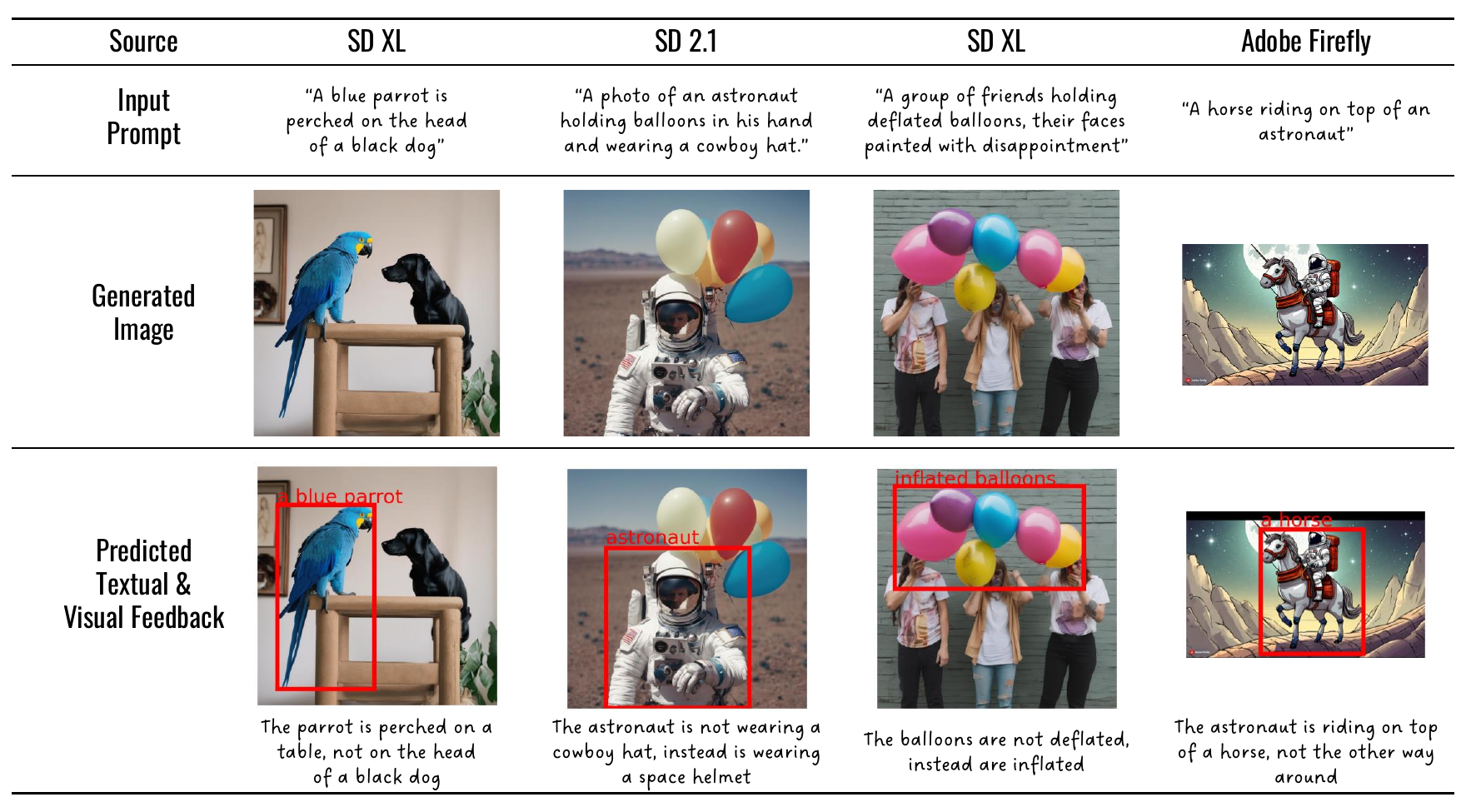}
\caption{Additional qualitative of out-of-distribution results using synthetic images generated using Stable-Diffusion XL~\cite{podell2023sdxl_stable_difussion}, StableDiffusion 2.1~\cite{rombach2021highresolution} and Adobe Firefly~\cite{Adobe_Firefly}.}
  \label{fig:supmat_ood}
\end{figure*}

\begin{figure}
  \centering
  \includegraphics[width=\columnwidth]{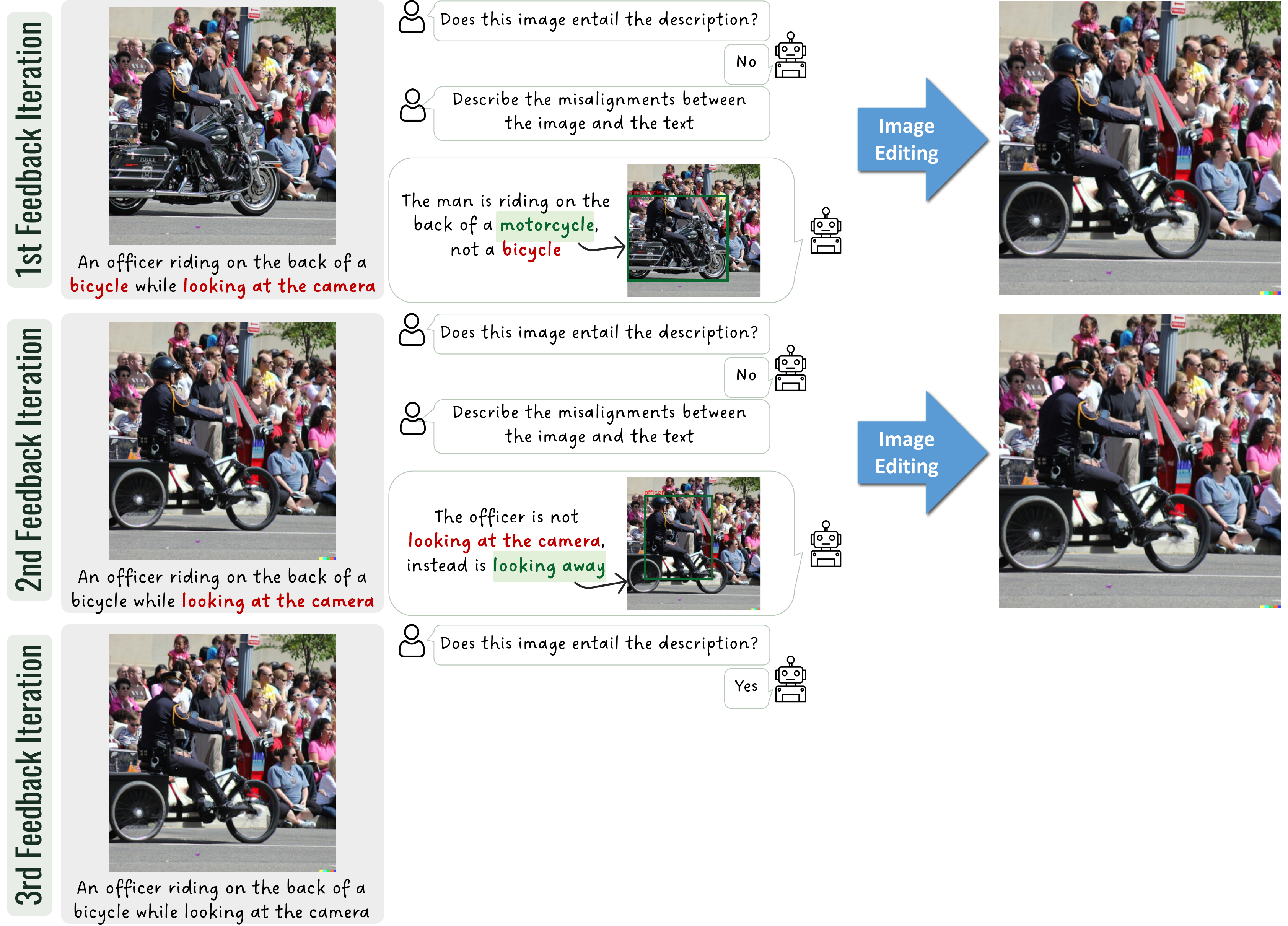}

\caption{
Demonstrating the iterative correction setup of image-text misalignments using a fine-tuned model on an example from the MagicBrush test dataset. In each iteration, the predicted feedback signals guide the model to sequentially correct the description until it fully aligns with the image.
}
\label{fig:serial_editing}

\end{figure}


\section{\method Details}
\label{supmat_section:method_details}
As detailed in the paper, to generate our training data using \method, we collected aligned text-image pairs from six different datasets: PickaPic~\cite{Kirstain2023PickaPicAO}, ImageReward~\cite{xu2023imagereward}, COCO~\cite{COCO}, Flickr30k~\cite{Plummer2015Flickr30kEC}, Open Images~\cite{openimages}, and ADE20k~\cite{ade20k_zhou2017scene}. It's important to note that we selected only the positive pairs from these datasets, totaling 1,302,201 text-image pairs. From these pairs, we generated negative examples with misalignments accompanied by textual and visual feedback. To validate the generation and maintain a high quality of generated data, we filtered out samples based on textual entailment tests, as detailed in \Cref{sec:generation_validation}, resulting in 1,087,912 text-image pairs for our filtered \method version.

\subsection{LLM Prompts}
\label{subsec:prompts}

As presented in Brown et al.~\cite{gpt3}, we also employ the few-shot Language Model (LLM) prompting technique. For each aligned image-caption pair, we generate a contradiction caption, a detailed explanation (feedback) of the misalignment between the original caption and its generated contradiction, a misalignment cue pointing to the contradiction source in the caption, and a textual label for the visual bounding box for the image. 

Our \method training dataset is compiled from six different datasets. To ensure the highest quality of generated data, we use a few-shot prompt for each misalignment type we want to generate, as well as per dataset, considering the varied caption styles across datasets. 

For images in COCO and Flickr30K with human-annotated captions, we choose the longest caption from multiple options, as it is usually the most descriptive. An example prompt for creating a \textit{Relation} type misalignment in the Coco dataset is in \Cref{fig:prompt_coco_relation}

The captions for PickaPic and ImageReward were predicted by the image captioning PaLI-2 model. For each image, five captions were generated, and we selected the caption with the highest score produced by the model. In \Cref{fig:prompt_pickapic_action}, we present the prompt used for the PickaPic dataset to generate an \textit{Action} misalignment.

ADE20k and OpenImages captions are taken from the Localized Narratives dataset. The dataset creators asked annotators to describe an image with their voice while simultaneously hovering their mouse over the region they are describing. This type of caption results in a small paragraph per image describing what can be seen in it, but not a short and concise caption summarizing the main idea in the image. Therefore, for these two datasets, we added another step before generating our labels. We generated a summarized caption from the image description and then applied the same procedure detailed before to this summarized caption. The prompt used for description summarizing is shown at \Cref{fig:prompt_ade20k_desc_to_caption}, and the prompt used for \textit{Attribute}
misalignment type for the ADE20k dataset is available at \Cref{fig:prompt_ade20k_noun}.

All our data is generated by PaLM-2 LLM API, the chosen parameters are: model=``chat-bison@001'', temperature=0.4, max-tokens=700,  top-p=0.95 and top-k=30. The entire prompts and data-generation pipeline will be released and available on our project page.

\subsection{Generation Validation}
\label{sec:generation_validation}
Although we strive to generate our data with the highest possible quality, there is still a potential for noise generation caused by LLM hallucination behavior or insufficient sample quality. To assess the quality of generated samples, we evaluate two crucial aspects: Contradiction Correctness and Feedback Correctness.

The Contradiction Correctness evaluation aims to assess whether the generated contradiction caption adequately contradicts the original one. Meanwhile, the Feedback Correctness evaluation checks whether the generated feedback accurately explains the misalignment between the original (aligned) image caption and the generated contradiction caption.

Textual Entailment~\cite{dagan2010recognizing} involves classifying whether a \emph{hypothesis} text is \emph{entailed} by a premise text. We employ an entailment model by Honovich~\etal~\cite{q2} to evaluate and filter our generated data. For the Contradiction Correctness evaluation, the input for our model includes the original (positive) caption as the premise and the generated contradiction as the hypothesis. We anticipate receiving a \emph{low} entailment score (NLI) in this evaluation, emphasizing that the negative caption truly differs from the original caption.

Feedback Correctness is evaluated by feeding the entailment model with the original image caption with the generated contradiction (premise) and the generated feedback (hypothesis). In this test, our goal is to receive a \emph{high} entailment score, indicating that the feedback remarks the misalignment between the original caption and the generated contradiction. A template for our evaluation tasks and examples of generated data with their evaluated entailment scores are shown in ~\Cref{fig:feedback_evaluation}.

Finally, we filter out from our initial dataset generated samples based on their evaluation. Samples with a Contradiction Correctness score higher than 0.25 and samples with a Feedback Correctness score lower than 0.75 are filtered out. In \Cref{fig:heatmap}, we present a heatmap of the remaining percentage of our generated dataset filtered by a range of thresholds. The selected thresholds for our released \method filtered version are $< 0.25$ for Contradiction Correctness score and $> 0.75$ for Feedback Correctness score.


\begin{figure}[t]
\centering
\resizebox{\columnwidth}{!}{%
    \begin{tikzpicture}
        \node[draw, rounded corners, fill=blue!10, align=left, text width=\columnwidth, inner sep=5pt, font=\scriptsize] {
        \underline{\textbf{Feedback Evaluation Task Template}}: \\
        \textbf{Contradiction Correctness:} \\
        \textbf{Premise:} $Original Caption$ \\
        \textbf{Hypothesis:} $Generated Contradiction$ \\
        \textbf{Feedback Correctness:} \\
        \textbf{Premise:} EXPECTED CAPTION: $Generated Contradiction$ . ACTUAL CAPTION: $Original Caption$ \\
        \textbf{Hypothesis:} $GeneratedFeedback$ \\
        };
    \end{tikzpicture}%
}
\resizebox{\columnwidth}{!}{%
    \begin{tikzpicture}
        \node[draw, rounded corners, fill=green!10, align=left, text width=\columnwidth, inner sep=5pt, font=\scriptsize] {
        \underline{\textbf{Example A}}: \\
        \textbf{Contradiction Correctness:} \\
        \textbf{Original Caption:} “A black cat playing on top of a wooden chair.”\\
        \textbf{Generated Contradiction:} “A black cat playing underneath a wooden chair”\\
        \textbf{Generated Feedback:} “The cat is on top of the chair, not underneath it.”\\
        \textbf{Contradiction Correctness score:} 0.02 \cmark \\
        \textbf{Feedback Correctness score:} 0.97 \cmark \\
        };
    \end{tikzpicture}%
}
\resizebox{\columnwidth}{!}{%
    \begin{tikzpicture}
        \node[draw, rounded corners, fill=red!10, align=left, text width=\columnwidth, inner sep=5pt, font=\scriptsize] {
        \underline{\textbf{Example B}}: \\
        \textbf{Contradiction Correctness:} \\
        \textbf{Original Caption:} “The man is riding along the beach in a two-wheeled cart pulled by a horse.”\\
        \textbf{Generated Contradiction:} “The man is driving along the beach in a two-wheeled cart  pulled by a horse.”\\
        \textbf{Generated Feedback:} “The man is riding along the beach, not driving”\\
        \textbf{Contradiction Correctness score:} 0.7 \xmark \\
        \textbf{Feedback Correctness score:} 0.95 \cmark\\
        };
    \end{tikzpicture}%
}
\resizebox{\columnwidth}{!}{%
    \begin{tikzpicture}
        \node[draw, rounded corners, fill=red!10, align=left, text width=\columnwidth, inner sep=5pt, font=\scriptsize] {
        \underline{\textbf{Example C}}: \\
        \textbf{Contradiction Correctness:} \\
        \textbf{Original Caption:} “a dining room table with a bottle of wine and wine glasses and a pot of carrots”\\
        \textbf{Generated Contradiction:} “a dining room table with a bottle of wine and wine glasses and a pot of potatoes”\\
        \textbf{Generated Feedback:} “The carrots are replaced with potatoes.”\\
        \textbf{Contradiction Correctness score:} 0.06 \cmark \\
        \textbf{Feedback Correctness score:} 0.01 \xmark \\
        };
    \end{tikzpicture}%
}

\caption{Automatic feedback evaluation. The blue sub-figure presents our evaluation template using the Honovich~\etal~\cite{q2} entailment model to calculate Contradiction Correctness and Feedback Correctness scores. The next three sub-figures are examples of generated data samples with their calculated scores. While example A (green sub-figures) passed our selected thresholds as good quality generated samples, examples B and C (red sub-figures) are filtered out. B is filtered out because its Contradiction Correctness score is higher than 0.25,  while C is filtered out because the received Feedback Correctness score is lower than 0.75.}
\label{fig:feedback_evaluation}
\end{figure}

\begin{figure*}
    \centering
    \includegraphics[width=\textwidth]{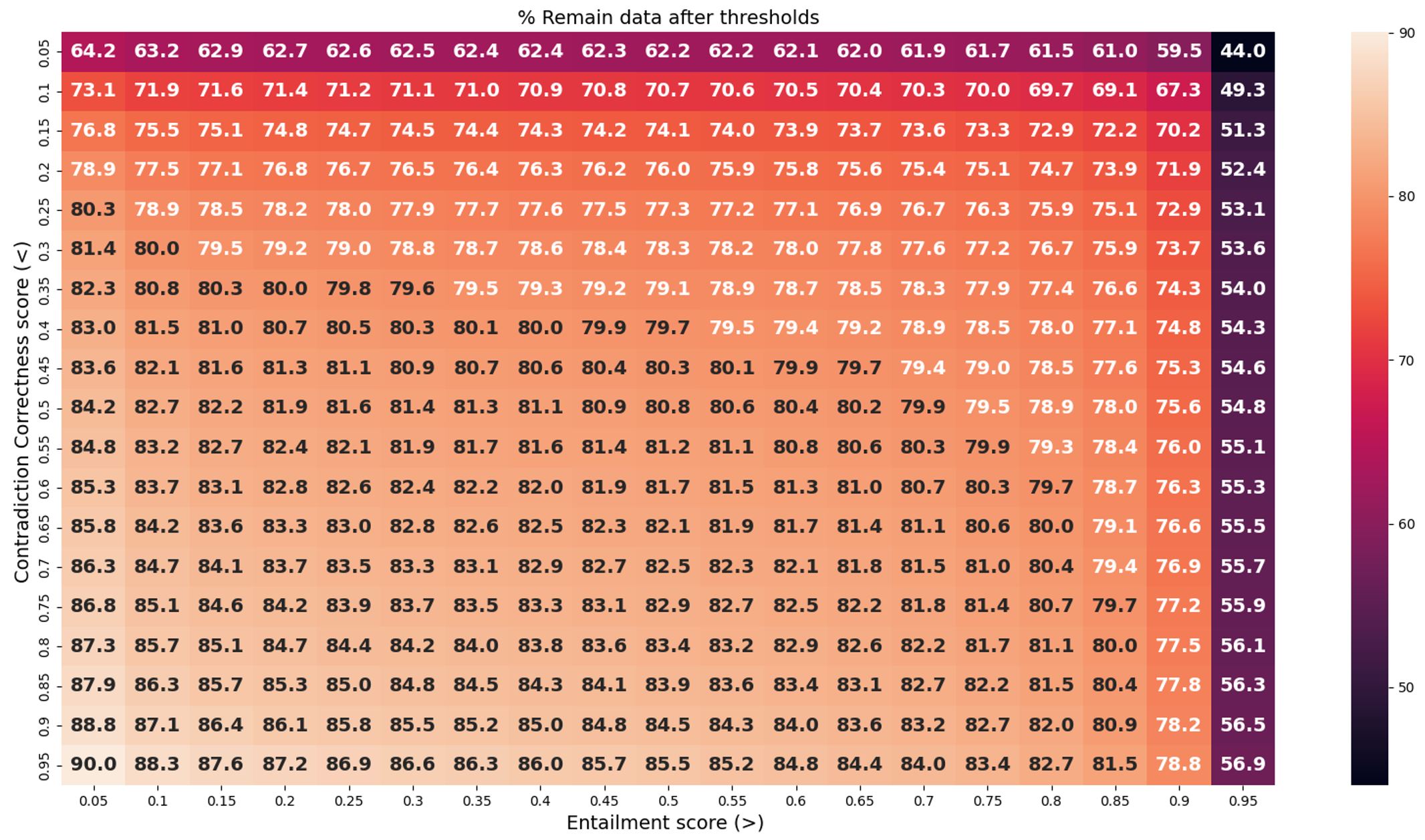}
\caption{A heatmap overlay presenting the percentage of remaining data from the entire set generated after applying a range of thresholds for Contradiction Correctness and Feedback Correctness scores.}
    \label{fig:heatmap}
\end{figure*}

\section{\testset Details}
Our \testset benchmark is constructed based on the SeeTRUE dataset~\cite{WYSIWYR}, which contains aligned and misaligned text-image pairs. Each misaligned pair is annotated with three human-generated misalignment feedbacks.

Similar to the approach detailed in \ref{subsec:prompts}, we utilize PaLM 2 to generate the required labels. In this case, there is no need to generate a contradiction and feedback for a sample. Instead, we need to summarize the three human feedbacks into a single and concise feedback and generate the remaining labels, such as the misalignment cue source in the caption and the textual label for the visual bounding box.

The few-shot prompt used for \testset generation is presented in \Cref{fig:prompt_seetrue}.

\section{Evaluation Metrics - Additional Details}
\label{supmat_section:evaluation_metrics}
This section builds upon \Cref{subsec:evaluation_metrics} by providing deeper and more technical details of the evaluation process and metrics calculation reported in \Cref{tab:main_results}. We compare our models against current leading VLM baselines using several metrics, which are explained in detail in this section. It is important to emphasize that our fine-tuned models deliver all required results in an end-to-end manner, whereas for the baselines, we must generate the answer using a different prompt each time.

To evaluate the quality of textual outputs, we conducted human annotation study as mentioned in \Cref{sec:experiments} and automatic metric calculation. \Cref{fig:ann_feedback_correlation} we show the annotators agreement for the feedback predictions correctness produced by each model. For the automatic metric to evaluate we opt for BART NLI~\cite{lewis2019bart}. Based on \Cref{fig:correlations}, we find it to be the most correlated metric with human ratings. The NLI algorithm receives two textual inputs: a `premise' and a `hypothesis', and outputs an entailment score ranging from 0 to 1, indicating semantic alignment (a higher score implies greater semantic alignment). In our experiments, we consider the ground truth text as the `premise' and the model prediction as the `hypothesis'. \Cref{fig:metrics_fig} in the main paper presents examples with metrics calculations results.

\textbf{Feedback NLI} (Textual Feedback Quality): To obtain feedback from the baseline models, we query them with the question: \textit{``What are the misalignments between the image and the text: <text>?''} As mentioned earlier, we use BART NLI to measure the semantic alignment between the ground truth and the predicted feedback answer.

\textbf{Textual Misalignment NLI}: In this task, we aim to guide the VLM to identify which part of the input caption causes misalignment between the image and the text. We pose the question to the model as follows: \textit{``Which part of <text> does not align with the image''} Again, we utilize the BART NLI model to assess the semantic quality of the predictions compared to the ground truth.

\textbf{Visual Misalignment - F1-Score@0.75}: This metric aims to assess the VLM's accuracy in marking the visual source of misalignment between the image and the caption. The visual misalignment is represented by a labeled bounding box. Since baseline VLMs cannot predict a bounding box, we divide this process into two steps. Initially, we query the VLM with: \textit{``Which part of the following image does not align with the <text>?''}. To obtain a bounding box corresponding to the predicted visual misalignment, we employ the GroundingDino~\cite{groundingdino} grounding model to extract a bounding box from the image, guided by the predicted label. Our fine-tuned models, trained with \trainset data, do not require this two-step process and can predict feedback, including bounding boxes, in a single inference step. In \Cref{tab:main_results}, we report the performance of our fine-tuned models using the same pipeline as the VLM baseline, as well as results from the PaLI models. The chosen metric for comparing model performance is the F1-Score@0.75, which sets the IoU threshold at 0.75 and balances precision and recall to measure the quality of localization.

\textbf{Binary Class Accuracy} (Image-Text Alignment): This metric compares the models' abilities to determine whether there is any misalignment between an image and its caption in a binary fashion. The input query for the VLMs is: \textit{``Does this image entail the description <text>?''}, expecting `yes' or `no' answers. The reported metric is the average accuracy across the SeeTRUE test dataset.

\begin{figure*}
  \centering
  \includegraphics[width=\textwidth]{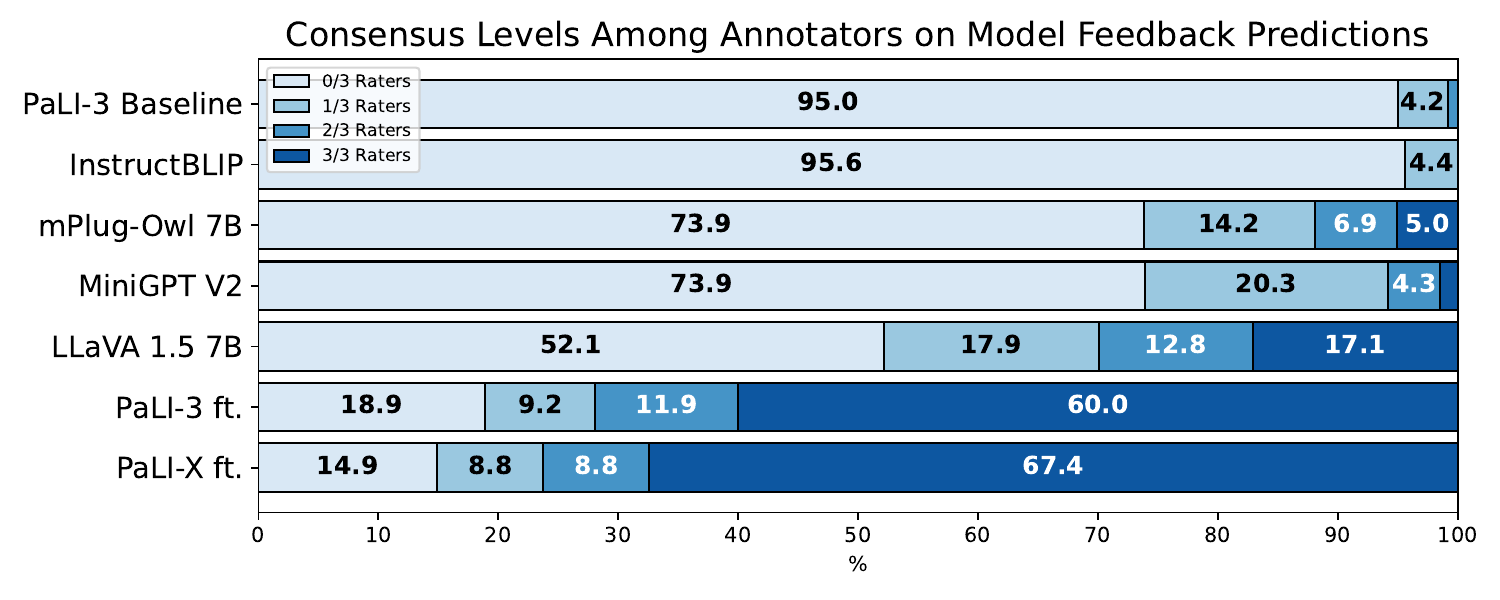}

\caption{
Bar chart showing consensus levels among three annotators for model feedback predictions, segmented by the number (0-3) of raters in agreement. For example, in the bottom bar, all three raters agreed on prediction's correctness in 67.4\% of cases, while for LLaVA, this occurred in only 17.1\% of cases.
}
  \label{fig:ann_feedback_correlation}
\vspace{-10pt}

\end{figure*}

\section{Training details}
\label{sec:reproducibility}

We provide details on the fine-tuning process for PaLI models. We fine-tune PaLI using T5X \cite{roberts2203scaling} on JAX \cite{bradbury2018jax}. The fine-tuning process involves utilizing the Adam optimizer with a learning rate of 2e-5. The model undergoes two epochs of training, and 10\% of the training set is reserved as a validation set for checkpoint selection.

For the supervised experiments, we fine-tuned the PaLI models with the visual question answering (VQA) task. This fine-tuning process aims to enhance the model's performance in specific tasks related to VQA, such as image-text alignment and providing textual and visual feedback.

The fine-tuning tasks include:

\begin{enumerate}
    \item \textbf{Image-Text Alignment:} In this task, we utilize the same query as in the zero-shot setup, \textit{``Does this image entail the description \texttt{<text>}?''}. The model is expected to provide a binary response, either `yes' or `no', indicating whether the image aligns with the given text description.

    \item \textbf{Textual and Visual Feedback:} This task involves introducing a query for combined feedback: \textit{``What are the misalignments between this image and the text \texttt{<text>}?''}. The expected response format comprises three components: feedback, misalignment in text, and misalignment in the image (bounding-box). The aim is to extract detailed feedback and specific misalignment indicators in a single model interaction.
\end{enumerate}

Hardware Requirements:
The computational resources required for fine-tuning vary depending on the specific PaLI model. PaLI and PaLI-3 fine-tuning require four v4 chips \cite{jouppi2020domain}, while fine-tuning for PaLI-X necessitates sixteen v4 chips.

\begin{figure*}
\promptfont
\centering
\begin{tabular}{p{\linewidth}}
\toprule
Context \\
\midrule
Your primary task is to generate subtly contradictory captions based on an original input caption. This test is intended to observe the capacity of models to recognize nuanced semantic changes which might be easily detected by humans but challenging for models.\\
\\
Guidelines:\\
Caption Modification: Introduce a delicate semantic change to the original caption. The alteration should subtly change the meaning while remaining potentially overlooked. It should neither be drastic nor too minor.\\
MISALIGNMENT: After constructing your contradictory caption, articulate the disparity between the original CAPTION and the CONTRADICTION in a crisp manner. The elucidation should:\\
Spotlight the deviation from the angle of the CONTRADICTION, pinpointing what the CONTRADICTION suggests but is countered by the CAPTION.\\
For the CAPTION description, ensure it thoroughly represents the true content of the CAPTION (e.g., ``CAPTION: a blue ball on a shelf'').\\
For the CONTRADICTION description, keep it brief and directly replicate the relevant segment (e.g., ``CONTRADICTION: green ball'').\\
MISALIGNMENT TYPE: Categorize the nature of the misalignment into one of the following categories: Object/Noun, Attribute/Adjective, Action/Verb, or Relation\\

\midrule
Few-Shot \\
\midrule

Guidelines:\\
- Relation Misalignment: Change the relationship or position between objects. This entails tweaking prepositions, positioning, or context which defines the relationship between the mentioned items. In the misalignment description, specifically indicate the changed relation and the affected subjects.- Mention at the SOURCE fields the changed preposition and related subject\\
\\
Examples: \\
\\
CAPTION: ``A laptop next to a stack of books on a wooden desk''\\
CONTRADICTION: A laptop under a stack of books on a wooden desk.\\
MISALIGNMENT: The laptop is next to the the books (CAPTION: laptop next to a stack of books), not under them (CONTRADICTION: laptop under a stack of books).\\
MISALIGNMENT TYPE: Relation\\
\\
CAPTION: Two cats are sitting below a branch of a tree in front of a building with windows and trees in the background .\\
CONTRADICTION: Two cats are sitting above a branch of a tree in front of a building with windows and trees in the background .\\
MISALIGNMENT: The cats are in front of the building sitting below a branch (CAPTION: Two cats sitting below a branch), not above it (CONTRADICTION: Two cats sitting above a branch)\\
MISALIGNMENT TYPE: Relation\\
\\
CAPTION: portrait of a redhead girl with blue eyes in a violet color t shirt sitting on a yellow color sofa near a gray wall\\
CONTRADICTION: portrait of a redhead girl with blue eyes in a violet color t shirt sitting on a yellow color sofa apart from a gray wall\\
MISALIGNMENT: The sofa is near the wall (CAPTION: a yellow sofa near a gray wall), not apart from it (CONTRADICTION: a yellow sofa apart from a gray wall)\\
MISALIGNMENT TYPE: Relation\\
\\
CAPTION: Old fashioned chests are stacked atop each other on a city block.\\
CONTRADICTION: Old fashioned chests are stacked next to each other on a city block.\\
MISALIGNMENT: The chests are stacked atop each other (CAPTION: chests stacked atop each other), not next to each other (CONTRADICTION: chests stacked next to each other).\\
MISALIGNMENT TYPE: Relation\\
\\
CAPTION: Boy holding large stuffed toy bear outside of building.\\
CONTRADICTION: Boy holding large stuffed toy bear inside of building.\\
MISALIGNMENT: The boy is outside of a building (CAPTION: Boy holding large stuffed toy bear outside of building), not inside of it  (CONTRADICTION: boy holding large stuffed toy bear inside of building) .\\
MISALIGNMENT TYPE: Relation\\
\\
Create a MISALIGNMENT of type: Relation\\
Remember when writing the MISALIGNMENT that the expectation is that CONTRADICTION should happen, but instead we got the scene of CAPTION. We expect for CONTRADICTION but CAPTION counters it.\\
CAPTION: A crystal bowl filled with oranges on top of a table. \\

\midrule
Generated \\
\midrule
CONTRADICTION: A crystal bowl filled with oranges beneath a table.\\
MISALIGNMENT: The bowl is on top of the table (CAPTION: bowl on top of a table), not beneath it (CONTRADICTION: bowl beneath a table).\\
MISALIGNMENT TYPE: Relation\\

\bottomrule
\end{tabular}   
\caption{Few-shot prompt example for generating a \emph{Relation} misalignment in the COCO~\cite{COCO} dataset.}
\label{fig:prompt_coco_relation}

\end{figure*}

\begin{figure*}
\promptfont
\centering
\begin{tabular}{p{\linewidth}}
\toprule
Context \\
\midrule
Your primary task is to generate subtly contradictory captions based on an original input caption. This test is intended to observe the capacity of models to recognize nuanced semantic changes which might be easily detected by humans but challenging for models.\\
Guidelines:\\
Caption Modification: Introduce a delicate semantic change to the original caption. The alteration should subtly change the meaning while remaining potentially overlooked. It should neither be drastic nor too minor.\\
MISALIGNMENT: After constructing your contradictory caption, articulate the disparity between the original CAPTION and the CONTRADICTION in a crisp manner. \\
The elucidation should:\\
Spotlight the deviation from the angle of the CONTRADICTION, pinpointing what the CONTRADICTION suggests but is countered by the CAPTION.\\
For the CAPTION description, ensure it thoroughly represents the true content of the CAPTION (e.g., "CAPTION: a blue ball on a shelf").\\
For the CONTRADICTION description, keep it brief and directly replicate the relevant segment (e.g., "CONTRADICTION: green ball").\\
MISALIGNMENT TYPE: Categorize the nature of the misalignment into one of the following categories: Object/Noun, Attribute/Adjective, Action/Verb, or Relation\\
\midrule
Few-Shot \\
\midrule

Guidelines:\\
- Action/Verb Misalignment: Modify a central action or verb in the original caption. This change should introduce a contradiction in terms of the activity or behavior described. The changed verb should make sense within the context and should neither be too drastic nor too obscure. Ensure the difference between the original caption's action and the contradictory action is clearly and succinctly articulated in the misalignment description.\\
\\
Examples: \\
\\
CAPTION: "A person dressed as Captain America is holding a shield in front of a wall."\\
CONTRADICTION: A person dressed as Captain America is throwing a shield and standing in front of a wall.\\
MISALIGNMENT: The person is not throwing a shield (CONTRADICTION: person throwing a shield), instead is holding it (CAPTION: person holding a shield)\\
MISALIGNMENT TYPE: Action/Verb\\
\\
CAPTION: "a close up of a lion with a long mane sitting on the grass with trees in the background"\\
CONTRADICTION: a close up of a lion with a long mane drinking water on the grass in the forest with trees in the background\\
MISALIGNMENT: The lion is sitting (CAPTION: a lion sitting), not jumping on the grass (CONTRADICTION: lion drinking water)\\
MISALIGNMENT TYPE: Action/Verb\\
\\
CAPTION: "A painting of a man standing in front of a fire in a field with trees and sky in the background ."\\
CONTRADICTION: A painting of a man running in front of a fire in a field with trees and a cloudy sky .\\
MISALIGNMENT: The main the painting is standing (CAPTION: man standing in front of a fire), not running (CONTRADICTION: man running in front of a fire)\\
MISALIGNMENT TYPE: Action/Verb\\
\\
CAPTION: "A tattooed woman in a purple brassiere and purple underpants is standing in front of a table with a bottle on it."\\
CONTRADICTION: A tattooed woman in a purple brassiere and purple underpants is dancing in front of a table with a bottle on it.\\
MISALIGNMENT: The woman is not dancing (CONTRADICTION: tattooed woman dancing), instead is sitting (CAPTION: A tattooed woman is standing)\\
MISALIGNMENT TYPE: Action/Verb\\
\\
CAPTION: "portrait of a smiling man and woman hugging each other on the street with trees and buildings in the background"\\
CONTRADICTION: portrait of a smiling man and woman fighting on the street with trees and buildings in the background\\
MISALIGNMENT: The man and woman are not fighting each other (CONTRADICTION: man and woman fighting), instead are hugging (CAPTION: a smiling man and woman hugging each other)\\
MISALIGNMENT TYPE: Action/Verb\\
\\
CAPTION: A black background with a picture of a guy in a red hat jumping with a skate board.\\
CONTRADICTION: A black background with a picture of a guy in a red hat drinking water with a skate board.\\
MISALIGNMENT: The guy is not drinking water (CONTRADICTION: guy drinking water), instead is sitting (CAPTION: guy sitting)\\
MISALIGNMENT TYPE: Action/Verb\\
\\
Create a MISALIGNMENT of type: Action/Verb\\
Remember when writing the MISALIGNMENT that the expectation is that CONTRADICTION should happen, but instead we got the scene of CAPTION. We expect for CONTRADICTION but CAPTION counters it.\\
CAPTION: A white cat is sitting on the grass in front of a house with trees and a blue sky in the background \\

\midrule
Generated \\
\midrule
CAPTION: A white cat is sitting on the grass in front of a house with trees and a blue sky in the background .\\
CONTRADICTION: A white cat is running on the grass in front of a house with trees and a blue sky in the background.\\
MISALIGNMENT: The cat is not running (CONTRADICTION: cat running), instead is sitting on the grass (CAPTION: cat sitting)\\
MISALIGNMENT TYPE: Action/Verb\\

\\

\bottomrule
\end{tabular}   
\caption{Few-shot prompt example for generating an \emph{Action} misalignment in the PickaPic~\cite{Kirstain2023PickaPicAO} dataset.}
\label{fig:prompt_pickapic_action}

\end{figure*}

\begin{figure*}
\centering
\promptfont
\begin{tabular}{p{\linewidth}}
\toprule
Context \\
\midrule

Task Definition: Extracting Concise Image Captions\\
\\
Input: \\
DESCRIPTION: A long description of an image containing various objects, people, and scene details.\\
\\
Output: \\
CAPTION: A single-sentence caption that preserve all the information from the description.\\
\\
Guidelines:\\
- The goal is to create a concise caption that effectively communicates the image's information written as a caption not a description.\\
- Include what people are doing and wearing from the description.\\
- Describe where objects are and how they look from the description.\\

\midrule
Few-Shot \\
\midrule

Examples:\\
\\
DESCRIPTION: Front we can see ball. Background it is blur. We can see trees, people and sky.\\
CAPTION: A ball with a blur background that includes trees, people and sky\\
\\
DESCRIPTION: There is a poster in which, there is a image. In the image, there are two vehicles parked on the grass on the ground. In the background, there are trees. Below this image, there are texts and watermark on the white page. \\
CAPTION: An image of two vehicles parked on the grass, below the image there is a text.\\
\\
DESCRIPTION: In this picture I can see food items, fork and a bowl with a liquid in it, on the plate, on an object.\\
CAPTION: Food items, fork and a bowl with liquid in it are on a plate.\\
\\
DESCRIPTION: This image is taken outdoors. At the top of the image there is the sky. At the bottom of the image there is a runway. In the background there are a few buildings and there is a ground with grass on it. In the middle of the image there are two airplanes on the runway and there is a pole. There are two objects on the runway. \\
CAPTION: Two airplanes on an outdoors runway with buildings and sky on the background\\
\\
DESCRIPTION: Here in this picture we can see an air plane present in an airport over there and we can also see a couple of trucks present beside it over there and we can see people standing on the ground over there and in the far we can see trees and plants present over there.\\
CAPTION: An airplane in an airport with trucks beside it, there are people standing on the ground.\\
\\
DESCRIPTION: In this picture there is a view of the airport. In the front there is a white color tower and some flights parked on the ground. In the background there are some buildings. On the top we can see the plane flying in the sky.\\
CAPTION: An airport view, at the foreground a white color tower and planes parked on the ground. Building are present in the background and a plane is flying in the sky\\

DESCRIPTION: This picture is a black and white image. In this image we can see one woman with black sunglasses walking, some plants near the building, one big pole, some text on the bottom left side corner of the image, some grass on the ground, one object and one wire attached to the pole.\\
CAPTION: A black and white image of a woman walking with black sunglasses, a building is present also with plants near to it\\

DESCRIPTION: In this picture I can observe a man and woman. Woman is holding a paper in her hand. In the background I can observe a building and plants.\\
CAPTION: A man with a woman who is holding a paper in her hand, there is a building and plants at the background.\\
\\
DESCRIPTION: In this image I can see few women sitting on the concrete wall and I can see a dog which is brown, cream and black in color is sitting on the ground. In the background I can see a person standing, the white colored cloth, few plants and a table, and on the table I can see a white colored cloth.\\
CAPTION: Women sitting on the concrete wall outside a building with a brown, cream and black dog. At the background there is plants, a table and a person wearing white colored cloth\\
\\
DESCRIPTION: In this picture we can see a flower vase and a name board on the platform and here we can see four people are standing on the floor. In the background we can see the name on the wall and we can see plants, roof and lights.\\
CAPTION: \\

\midrule
Generated \\
\midrule
CAPTION: People standing on the floor near a flower vase and a name board.\\
\bottomrule
\end{tabular}   
\label{fig:prompt_ade20k_desc_to_caption}

\caption{Few-shot prompt example for generating a short and concise caption from a lengthy image description annotation for the Localized Narratives~\cite{pont2020connecting} datasets: ADE20k~\cite{ade20k_zhou2017scene} and OpenImages~\cite{openimages}.}

\end{figure*}

\begin{figure*}
\centering
\promptfont
\begin{tabular}{p{\linewidth}}
\toprule
Context \\
\midrule
Your primary task is to generate subtly contradictory captions based on an original input caption. This test is intended to observe the capacity of models to recognize nuanced semantic changes which might be easily detected by humans but challenging for models.\\
\\
Guidelines:\\
Caption Modification: Introduce a delicate semantic change to the original caption. The alteration should subtly change the meaning while remaining potentially overlooked. It should neither be drastic nor too minor.\\
MISALIGNMENT: After constructing your contradictory caption, articulate the disparity between the original CAPTION and the CONTRADICTION in a crisp manner. The elucidation should:\\
Spotlight the deviation from the angle of the CONTRADICTION, pinpointing what the CONTRADICTION suggests but is countered by the CAPTION.\\
For the CAPTION description, ensure it thoroughly represents the true content of the CAPTION (e.g., "CAPTION: a blue ball on a shelf").\\
For the CONTRADICTION description, keep it brief and directly replicate the relevant segment (e.g., "CONTRADICTION: green ball").\\
MISALIGNMENT TYPE: Categorize the nature of the misalignment into one of the following categories: Object/Noun, Attribute/Adjective, Action/Verb, or Relation\\

\midrule
Few-Shot \\
\midrule

Guidelines:\\
- Attribute/Adjective Misalignment: Alter an attribute or adjective that describes an object or scene in the original caption. This alteration should create a contradiction in the description. Ensure that the changed attribute/adjective remains plausible within the context of the caption and doesn't alter the core meaning. The discrepancy between the original caption's attribute and the contradictory attribute should be highlighted in the misalignment description.\\
\\
Examples: \\
\\
CAPTION: "a waterfall in a tropical forest with green plants and trees"\\
CONTRADICTION: a waterfall in a snowy forest with green plants and trees\\
MISALIGNMENT: The forest is not snowy (CONTRADICTION: snowy forest), instead is a tropical one (CAPTION: tropical forest)\\
MISALIGNMENT TYPE: Attribute/Adjective\\
\\
CAPTION: "a cartoon of a red haired mermaid holding a fish in the sea"\\
CONTRADICTION: a cartoon of a mermaid with black hair in the water\\
MISALIGNMENT: The hair of the cartoon mermaid is red (CAPTION: red haired mermaid), not black (CONTRADICTION: mermaid with black hair)\\
MISALIGNMENT TYPE: Attribute/Adjective\\
\\
CAPTION: portrait of a naked woman with long hair on a grey background\\
CONTRADICTION: A portrait of a naked woman with short hair on a grey background\\
MISALIGNMENT: The woman has long hair (CAPTION: long hair), not short hair (CONTRADICTION: short hair)\\
MISALIGNMENT TYPE: Attribute/Adjective\\
\\
CAPTION: The colorful flowers are in the vase next to the window.\\
CONTRADICTION: The white and black flowers are in the vase next to the window.\\
MISALIGNMENT: The flowers are colorful (CAPTION: colorful flowers) , not white and black (CONTRADICTION: white and black flowers)\\
MISALIGNMENT TYPE: Attribute/Adjective\\
\\
CAPTION: Vase sitting next to the window with droopy flowers in it.\\
CONTRADICTION: Vase sitting next to the window with fresh flowers in it.\\
MISALIGNMENT: The vase has droppy flowers (CAPTION: droppy flowers), not fresh flowers (CAPTION: fresh flowers)\\
MISALIGNMENT TYPE: Attribute/Adjective\\
\\
CAPTION: A bicycle chained to a red pole and a green pole on the sidewalk in front of a building.\\
CONTRADICTION: A bicycle chained to a red pole and a blue pole on the sidewalk in front of a building.\\
MISALIGNMENT: One of the poles is red (CAPTION: green pole), not blue (CONTRADICTION: blue pole)\\
MISALIGNMENT TYPE: Attribute/Adjective\\
\\
Create a MISALIGNMENT of type: Attribute/Adjective\\
Remember when writing the MISALIGNMENT that the expectation is that CONTRADICTION should happen, but instead we got the scene of CAPTION. We expect for CONTRADICTION but CAPTION counters it.\\
CAPTION: A wall with light and dust particles under a wooden beam. \\

\midrule
Generated \\
\midrule
CONTRADICTION: A wall with light and dust particles under a metal beam.\\
MISALIGNMENT: The beam is wood (CAPTION: wooden beam), not metal (CONTRADICTION: metal beam)\\
MISALIGNMENT TYPE: Attribute/Adjective\\
\bottomrule
\end{tabular}   
\caption{Few-shot prompt example for generating an \emph{Attribute} misalignment in the ADE20k~\cite{ade20k_zhou2017scene} dataset.}
\label{fig:prompt_ade20k_noun}

\end{figure*}

\begin{figure*}
    \centering
    \promptfont

    \begin{tabular}{p{\linewidth}}
    \toprule
Context\\
\midrule
Your primary task is to generate a single summarized feedback based on visual feedback provided in the input, and identify the source of misalignment between the caption and the visual content. The goal is to extract meaningful insights from the feedback and link them to the corresponding elements in the caption.\\

Guidelines:\\
SUMMARIZED FEEDBACK: Given a caption and a set of visual feedbacks, your objective is to create a concise and informative feedback statement that captures the most significant misalignment between the caption and the visual content. No need to mention "at the caption" or "at the image".\\
Misalignment Identification: After constructing the summarized feedback, you should indicate the source of misalignment for both the caption and the visual content. In parentheses, specify the origin of the misalignment as follows:\\
CAPTION: Describe the source of misalignment that originates from the text (e.g., a description or detail in the caption that does not match the image). Mention also the related subject.\\
CONTRADICTION: Based on the feedback, pinpoint the source of misalignment in the visual content that contradicts the caption. Make sure the CONTRADICTION refers only to the relative part in FEEDBACKS. Mention also the related subject.\\
\midrule
Few-Shot \\
\midrule
Examples:\\
\\
CAPTION: "A woman with a blue shirt and yellow flower headband, sitting on a wooden bench outside on the grass."\\
FEEDBACKS: ["The woman's shirt is yellow, not blue.", "The woman sitting on the wooden bench is wearing a white shirt, not a blue shirt.", "The woman has a yellow shirt on, not blue."]\\
MISALIGNMENT: The sitting woman is wearing a yellow shirt (CONTRADICTION: yellow shirt), not a blue shirt (CAPTION: blue shirt)\\
\\
CAPTION: A couch on the left of a chair.\\
FEEDBACKS: ["A couch is to the left of a coffee table, there is no chair present.", "A couch and two tables under a window, there is no chair.", NaN]\\
MISALIGNMENT: The couch is to the left of a table (CONTRADICTION: coffe table), not left of a chair (CAPTION: chair).\\
\\
CAPTION: A sheep to the right of a wine glass.\\
FEEDBACKS: ["A sheep is facing the camera, not to the right of a wine glass, which is not present.", "A sheep looking at the camera, there is no wine glass.", "There is only a sheep alone in a field, it is not to the right of a wine glass"]\\
MISALIGNMENT: There is no wine glass (CAPTION: wine glass) (CONTRADICTION: no wine glass)\\
\\
CAPTION: A tomato has been put on top of a pumpkin on a kitchen stool. There is a fork sticking into the pumpkin. The scene is viewed from above.\\
FEEDBACKS: [NaN, "A pumpkin has been put on top of a kitchen stool. There is a fork sticking into the pumpkin. The scene is viewed from above. There is no tomato on top of the pumpkin.", "There is only a pumpkin, and no tomato"]\\
MISALIGNMENT: There is no tomato on top of a pumpkin (CAPTION: tomato on top of a pumpkin) (CONTRADICTION: no tomato)\\
\\
CAPTION: a red glass ball with a reflection on top of a gray surface\\
FEEDBACKS: ["The glass ball is blue, not red.", "The ball with the reflection should be blue instead of red.", "The ball is blue, not red"]\\
MISALIGNMENT: The color of the ball is blue (CONTRADICTION: blue ball), not red (CAPTION: red ball)\\
\\
CAPTION: A lunch salad in a yellow bowl made out of fruit, vegetables, and meats with chopsticks. \\
FEEDBACKS: ["There is no meat present, nor are there any chopsticks, simply the bowl with fruit and vegetable.", "There is a bowl of food, but no chopsticks.", "There is a salad with fruit and vegetables in a yellow bowl but there does not appear to be meat in it and there are no chopsticks."]\\
MISALIGNMENT: There is a yellow bowl with fruit and vegetables (CONTRADICTION: salad with fruit and vegetables in a yellow bowl) but without meat, also no chopsticks present (CAPTION: salad made out of fruit, vegetables, and meats with chopsticks)\\
\\
CAPTION: A cup filled with umbrellas and canes next to a white wall.\\
FEEDBACKS: ["The wall is gray, not white.", "A bucket has two canes in it, not umbrellas and canes and it is front of a gray wall, not a white wall.", "A cup has canes in it, but no umbrellas"]\\
MISALIGNMENT: The cup has canes in it (CAPTION: umbrellas and canes), but no umbrellas (CONTRADICTION: no umbrellas). The wall is not white (CAPTION: white tall), instead is gray (CONTRADICTION: gray wall).\\
\\
CAPTION: Small white toilet with spare rolls of toilet paper beside it.\\
FEEDBACKS:["A small white toilet is here without any rolls of toilet paper, not any spares.", "There is a toilet but no toilet paper.", "The toilet only has curtains/walls next to it, not spare rolls of toilet paper."]\\
MISALIGNMENT: The toilet does not have any spare rolls of toilet paper beside it (CONTRADICTION: NO toilet paper), as described in the caption (CAPTION: spare rolls of toilet paper)\\
\\
CAPTION: A cat is holding a frisbee in its mouth\\
FEEDBACKS: ["A dog is holding a frisbee in its mouth", "A dog is holding a frisbee in its mouth.", "A dog is holding a frisbee, not a cat"]
\\
MISALIGNMENT: \\

\midrule
Generation\\
\midrule
CAPTION: A cat is holding a frisbee in its mouth\\
FEEDBACKS: ["A dog is holding a frisbee in its mouth", "A dog is holding a frisbee in its mouth.", "A dog is holding a frisbee, not a cat"]\\
MISALIGNMENT: The animal holding the frisbee is a dog (CONTRADICTION: dog holding a frisbee), not a cat (CAPTION: cat holding a frisbee)\\
\bottomrule
\end{tabular}    
\caption{Few-shot prompt example for generating the required labels for our \testset test dataset from human-annotated feedbacks.}
\label{fig:prompt_seetrue}

\end{figure*}

\fi

\end{document}